\documentclass[10pt,twocolumn,letterpaper]{article}

\usepackage{iccv}
\usepackage{times}
\usepackage{epsfig}
\usepackage{epstopdf}
\usepackage{graphicx}
\usepackage{amsmath}
\usepackage{amssymb}
\usepackage{booktabs}
\usepackage{multirow} 
\usepackage{subcaption}
\usepackage{overpic}
\usepackage{color,xcolor}
\usepackage{amsmath}
\usepackage{marvosym}
\usepackage{balance}

\usepackage[breaklinks=true,bookmarks=false, colorlinks=true]{hyperref}

\iccvfinalcopy 

\def\iccvPaperID{} 

\ificcvfinal\fi

\begin{document}

\title{Random Sub-Samples Generation for Self-Supervised Real Image Denoising}

\author{Yizhong Pan\textsuperscript{1}, Xiao Liu\textsuperscript{1}, Xiangyu Liao\textsuperscript{1}, Yuanzhouhan Cao\textsuperscript{2}, Chao Ren\textsuperscript{1(\Letter)} \\
\textsuperscript{1}College of Electronics and Information Engineering, Sichuan University, China\\
\textsuperscript{2}School of Computer and Information Technology, Beijing Jiaotong University, China\\
{\tt\small \{panyizhong, liux, liaoxiangyu1\}@stu.scu.edu.cn, yzhcao@bjtu.edu.cn, chaoren@scu.edu.cn}
}

\maketitle
\ificcvfinal\thispagestyle{empty}\fi

\renewcommand{\thefootnote}{\Letter}
\footnotetext{Corresponding Author}

\begin{abstract}
	With sufficient paired training samples, the supervised deep learning methods have attracted much attention in image denoising because of their superior performance. However, it is still very challenging to widely utilize the supervised methods in real cases due to the lack of paired noisy-clean images. Meanwhile, most self-supervised denoising methods are ineffective as well when applied to the real-world denoising tasks because of their strict assumptions in applications. For example, as a typical method for self-supervised denoising, the original blind spot network (BSN) assumes that the noise is pixel-wise independent, which is much different from the real cases. To solve this problem, we propose a novel self-supervised real image denoising framework named Sampling Difference As Perturbation (SDAP) based on Random Sub-samples Generation (RSG) with a cyclic sample difference loss. Specifically, we dig deeper into the properties of BSN to make it more suitable for real noise. Surprisingly, we find that adding an appropriate perturbation to the training images can effectively improve the performance of BSN. Further, we propose that the sampling difference can be considered as perturbation to achieve better results. Finally we propose a new BSN framework in combination with our RSG strategy. The results show that it significantly outperforms other state-of-the-art self-supervised denoising methods on real-world datasets. The code is available at \href{https://github.com/p1y2z3/SDAP}{https://github.com/p1y2z3/SDAP}.
\end{abstract}

\section{Introduction}
\label{sec:intro}

Image denoising is a hot topic in low-level vision tasks, aiming to obtain high-quality images from noisy versions. In recent years, learning-based methods have been widely used in image denoising with their superior performance. Thanks to the construction of paired image datasets, the supervised approaches \cite{zhang2017beyond,anwar2019real,ren2021adaptive,cheng2021nbnet,tu2022maxim,wang2022uformer} can be efficiently implemented. In these approaches, it is a common practice to synthesize paired datasets directly using a specific noise model, such as the signal-independent additive white Gaussian noise (AWGN) model. 

However, the real images are captured by cameras with image signal processing (ISP) pipelines \cite{brooks2019unprocessing}. The noise is usually signal-dependent and spatially correlated in the real world. Moreover, ISP includes many non-linear operations that can complicate the noise distribution. Due to the influence of ISP pipelines within the camera, the noise distribution of real images is difficult to predict. Therefore, it is difficult and challenging for us to model the noise for real images. To overcome the problem of directly and explicitly modelling noise distribution, paired real-world datasets have been constructed by researchers. However, collecting paired noisy-clean images is difficult and time-expensive.   

\begin{figure}
	
	\centering
	\begin{subfigure}{0.32\linewidth}
		\centering
		\includegraphics[width=1in]{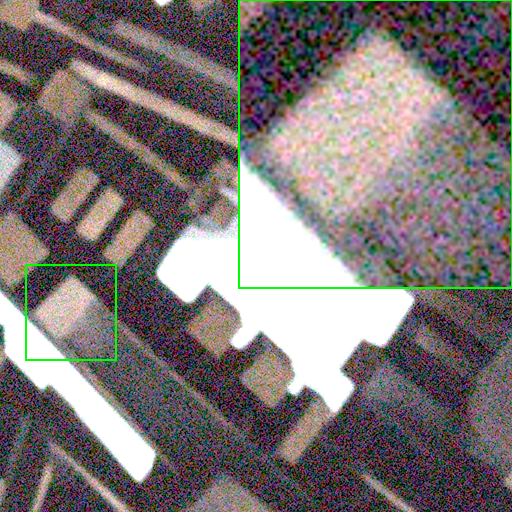}
		\scriptsize{Noisy}
	\end{subfigure}
	\begin{subfigure}{0.32\linewidth}
		\centering
		\includegraphics[width=1in]{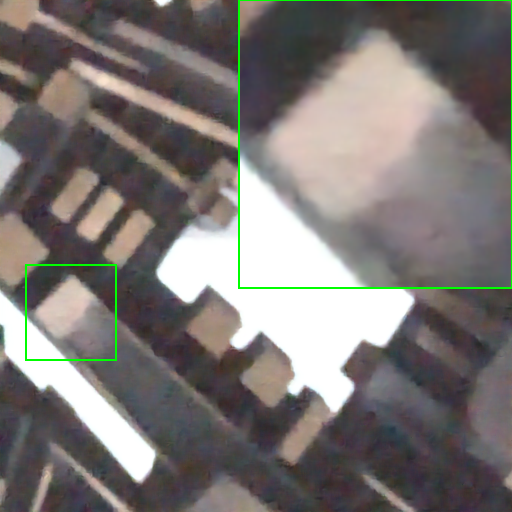}
		\scriptsize{CBDNet \cite{guo2019toward}}
	\end{subfigure}
	\begin{subfigure}{0.32\linewidth}
		\centering
		\includegraphics[width=1in]{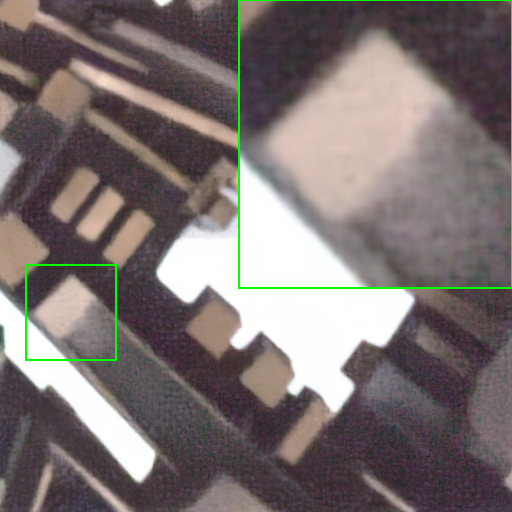}
		\scriptsize{C2N \cite{jang2021c2n}}
	\end{subfigure}
	\begin{subfigure}{0.32\linewidth}
		\centering
		\includegraphics[width=1in]{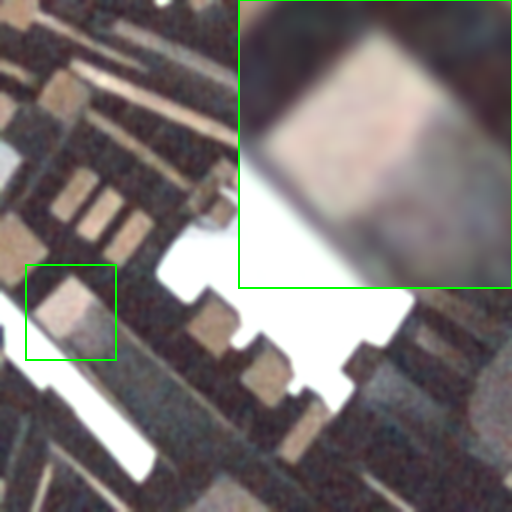}
		\scriptsize{CVF-SID \cite{neshatavar2022cvf}}
	\end{subfigure}
	\begin{subfigure}{0.32\linewidth}
		\centering
		\includegraphics[width=1in]{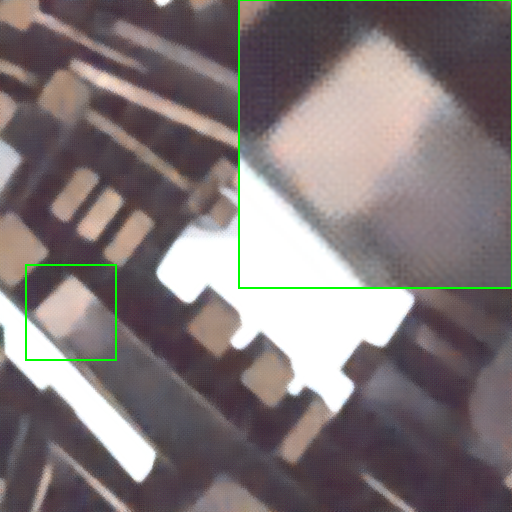}
		\scriptsize{AP-BSN \cite{lee2022ap}}
	\end{subfigure}
	\begin{subfigure}{0.32\linewidth}
		\centering
		\includegraphics[width=1in]{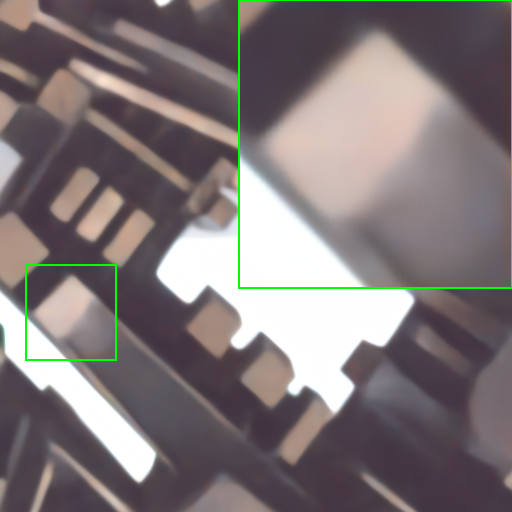}
		\scriptsize{\textbf{SDAP (S)(E) (Ours)}}  
	\end{subfigure}
	\vspace{-0.1cm}
	\caption{Real-world image denoising results on DND \cite{plotz2017benchmarking}.} 
	\vspace{-0.2cm}
	\label{dnd}
\end{figure}

Recently, a series of unsupervised and self-supervised methods have been introduced. These methods do not require paired datasets for training, and thus further avoiding the difficulty of collecting real-world paired datasets for denoising. Among them, Noise2Void \cite{krull2019noise2void} proposes Blind-Spot Network (BSN) that can be trained with only single noisy images. Unfortunately, BSN has strict assumptions, which can only be applied to pixel-wise independent noise. AP-BSN \cite{lee2022ap} applies pixel-shuffle downsampling (PD) strategy \cite{zhou2020awgn} to BSN, remove the spatial correlation in the real-world noise, and thus the real image can also meet the conditions for BSN to some extent. However, real noisy datasets usually have relatively limited numbers of samples. AP-BSN directly adopts the strategy of PD, which can only obtain $s^{2}$ sub-images for a noisy image (the stride factor of PD is $s$). This will inevitably lead to the insufficient samples, and make the training of BSN less effective. Therefore, the optimal performance of BSN cannot be achieved.

We observe that adding some perturbations to the training images can greatly expand the training data, which improves the performance of BSN.  Further, we consider the samples difference as the perturbation. To obtain more random perturbations and more sub-samples, we propose random sub-samples generation (RSG) strategy to break the fixed sampling pattern of PD. Based on this, we propose a new cyclic sampling difference loss for BSN and a new BSN framework in combination with our RSG strategy. In summary, our contributions include:
\begin{itemize}
	\item We provide a new idea about adding perturbations to the training data to improve the BSN performance. Then, we suggest that the sub-samples difference generated by sampling can be considered as perturbations for higher performance.
	
	\item We propose a new self-supervised framework for real image denoising with random sub-samples generation and cyclic sampling difference loss. 
	\item Our method performs very favorably against state-of-the-art  (SOTA) self-supervised denoising methods on real-world datasets.
\end{itemize}

\section{Related Work}
\label{sec:formatting}

The image denoising task aims to restore a clean image from its noisy counterpart. There are two main types of non-learning-based image denoising: filtering-based methods and model-based methods. Filter-based methods involve using some artificially designed low-pass filter to remove image noise. Taking advantage of the condition of having many similar image blocks in the same image, noise can also be removed by the local similarity of image, such as NLM \cite{buades2005non}, BM3D \cite{dabov2007image}, etc. Model-based methods involve modeling the distribution of natural images or noise and then using the model distribution before obtaining a clear image with an optimization algorithm, such as WNNM \cite{gu2014weighted}. The learning-based methods are effective ways to reduce image noise. Usually, it can be divided into traditional and deep network-based methods. In recent years, due to the continuous development of the deep learning technology, deep network-based methods have achieved superior denoising results compared to previous methods, and thus they have become the mainstream denoising methods. Generally, deep network-based methods can be further classified according to their training manners.

\subsection{Supervised Image Denoising}
Most of the early works \cite{mao2016image,zhang2017beyond,tai2017memnet,zhang2018ffdnet,zheng2021deep} assume that the noise is independent and uniformly distributed, and additive Gaussian white noise (AWGN) is usually used to model image noise and synthetic paired datasets. These methods achieve SOTA results on AWGN denoising tasks. However, the noise model in the real world is complex and unknown. Models trained with synthetic datasets are difficult to be applied to the real world. Recently, most of the methods \cite{guo2019toward,anwar2019real,kim2020transfer,yue2020dual,zamir2021multi,cai2021learning,zamir2022restormer,ren2022enhanced,chen2022simple} use produced real-world paired datasets for training. Due to the differences in camera parameters, these methods may not be widely used in the real world.

\subsection{Pseudo-Supervised Image denoising}
The supervised denoising methods require paired noisy-clean images, while the pseudo-supervised denoising methods relax the data requirements. Noise2Noise \cite{lehtinen2018noise2noise} proposes that pairs of noisy images can be used to train the network when the noise mean is zero. When some unpaired noisy-clean images are available, some methods \cite{chen2018image,cha2019gan2gan,hong2020end,jang2021c2n} learn the noise distribution of noisy images by generative adversarial networks (GAN). Then, clean images are used to generate pseudo-paired noisy-clean images to train the denoising network. However, these methods are still difficult to be applied because the scene distribution of noisy images often does not match the available clean images \cite{jang2021c2n}.

\subsection{Self-Supervised Image Denoising}
The self-supervised denoising methods require single noisy images for training. Noise2Void \cite{krull2019noise2void} finds that masking a portion of pixels in noisy images for pairing can train the denoising network, and thus proposes a BSN for self-supervised denoising. Laine19 \cite{laine2019high} and D-BSN \cite{wu2020unpaired} further optimize the BSN and improve its performance. Noise2Same \cite{xie2020noise2same} and Blind2Unblind \cite{wang2022blind2unblind} propose new denoising losses for self-supervised training through mask-based blind-spot methods. Neighbor2Neighbor \cite{huang2021neighbor2neighbor} samples the noisy image into two similar sub-images to form a noisy-noisy pair for training. CVF-SID \cite{neshatavar2022cvf} can disentangle the clean image, signal-dependent and signal-independent noises from the real-world noisy input via various self-supervised training objectives. AP-BSN \cite{lee2022ap} combines PD and BSN to process real-world sRGB noisy images and employs asymmetric PD stride factors for training and inference. However, current self-supervised methods for real noise removal usually still require a large number of noisy images and are challenging to train with less training data to achieve superior performance.

\section{Proposed Method}

\subsection{Revisiting Noise2Void}
Noise2Void (N2V) \cite{krull2019noise2void} requires only single noisy images for training, and it assumes that the noise $n$ is pixel-wise independent (the noises at different pixel positions are independent) given the clean image $x$. $ y $ is the noisy image, i.e. $y = x + n$. Due to the local correlation of the pixels within an image, clean central pixels can be estimated from the surrounding noisy pixels. Therefore, using the pixel of the noisy image as a target for its surrounding pixels will not obtain the noisy pixel. Thus, N2V can train the denoising network in a self-supervised approach. Specifically, N2V loss $L_{N2V}$ can be written as follows:
\begin{align}
	L_{N2V} = E_{y}  \{\left \| f(y_{r}) - y_{c} \right \|^{2}_{2}\},
\end{align}
where $f(\cdot)$ is the normal denoising network, $y_{r}$ is the patch of $ y $ centered on $y_{c}$ with the same dimensions as the network receptive field. Note that $y_{r}$ does not contain $y_{c}$.

To facilitate the implementation of training with $L_{N2V}$, N2V proposes BSN $B(\cdot)$ capable of masking centroids. The noisy images used to train the BSN must satisfy the BSN assumption that the noise is pixel-wise independent and zero-mean. 
BSN can be trained by optimizing the following loss $L_{BSN}$:
\begin{align}
	L_{BSN} = E_{y} \{\left \| B(y) - y \right \|^{2}_{2}\}.
	\label{BSN}
\end{align}

$L_{N2V}$ and $L_{BSN}$ are equivalent for training BSN. For a known noise model (e.g. AWGN model), we can synthesize an infinite number of different noisy images. Thus, BSN can be easily trained. In contrast, since real image datasets often have a limited number of captured images, it is difficult to train BSN directly for real image denoising. Moreover, real images do not satisfy the pixel-wise independent noise assumptions of BSN. Therefore, the limitations of $L_{BSN}$ are significant, and it is not appropriate to use real images directly for BSN.

\subsection{AP-BSN and Its Limitation}
To make BSN better for the spatially correlated real image noise, AP-BSN \cite{lee2022ap} introduces PD \cite{zhou2020awgn} to break the spatial correlation among real noisy pixels, leading to near pixel-wise independent noise. Therefore, the real image after PD can be considered to meet the BSN assumption \cite{lee2022ap}, and thus can be used as the input for BSN. The original AP-BSN loss is written as $\left \| P\!D^{-1}(B(P\!D(y))) - y \right \|_{1}$, where $P\!D$ is the PD operator and $P\!D^{-1}$ is the inverse operator. To facilitate the analysis, we further write it as:
\begin{align}
	L_{AP\!-\!BSN} = E_{y} \{\left \| B(P\!D(y)) - P\!D(y) \right \|_{1}\}.
	\label{APBSN}
\end{align}

It can be observed that the pattern of pairing is fixed in Eq. (\ref{APBSN}) (i.e. fixed $P\!D(y)$ to $P\!D(y)$). We consider that the training of fixed patterns does not affect the performance of BSN when the sample number is large enough. In contrast, it is well known that insufficient sample number may lead to overfitting \cite{ying2019overview}. Overfitting of noisy-clean pairing training can only have a superior effect on the training datasets and fail on other datasets. However, BSN requires noisy-noisy pairs for training. The overfitting of BSN will make the denoising result fit the noisy image and affect the denoising performance. For synthetic noise, because the noise modelling can generate an infinite number of noisy images, BSN can avoid fitting the noise, resulting in a clean image. But when BSN is used for real noise, real datasets tend to have a much smaller sample number than synthetic datasets. In this case, BSN may fail to get high-quality clean pixels. 

To verify the effect of the number of training images on the denoising results, we train BSN with synthetic images. In this way, we can synthesize any number of noisy images. We employ DIV2K dataset \cite{timofte2017ntire} and add Gaussian noise with a level $\sigma = 25$  to synthesize noisy images. Two ways are adopted to train BSN with $L_{BSN}$: (1) adding noise with fixed seed in each epoch, and (2) adding noise with random seed in each epoch. Then, we perform denoising experiments on the test datasets generated with Gaussian noise (Set12 \cite{zhang2017beyond}, BSD68 \cite{roth2009fields}, and Urban100 \cite{huang2015single}). Table \ref{overfitting} shows the PSNR(dB) and SSIM results in the test datasets for different training strategies. We observe that when a limited number of noisy images (noisy images with fixed seeds) are used to train the BSN , the denoising performance degrades significantly.

Therefore, the introduction of PD is practical. Still, considering that the training data of real images are often limited, the direct use of $L_{AP\!-\!BSN}$ may affect the denoising performance for real image denoising.

\begin{table}\small 
	\centering
	\setlength{\tabcolsep}{1mm}
	\begin{tabular}{cccc}
		\toprule
		Dataset & Set12 & BSD68 & Urban100\\
		\midrule
		Fixed seed & 28.91/0.8192 &  27.26/0.7477 & 27.66/0.8172\\
		Random seed & 29.36/0.8399 &  27.72/0.7728 & 28.81/0.8632\\
		\bottomrule
		\vspace{-0.4cm}
	\end{tabular}
	\caption{The effect of different BSN training strategies on PSNR (dB)/SSIM. }
	\label{overfitting}
	\vspace{-0.3cm}
\end{table}

\begin{figure*}
	\centering
	\begin{subfigure}{0.32\linewidth}
		\centering
		\includegraphics[width=2.3in]{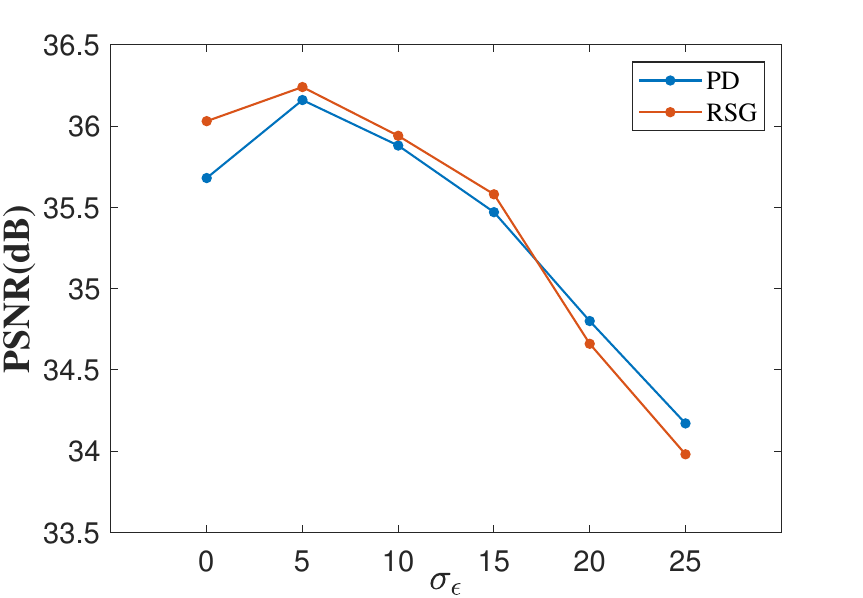}
		\vspace{-0.6cm}
		\caption{$L_{PBSN_{1}}$}
		\label{PBSN1}
	\end{subfigure}
	\hfill
	\begin{subfigure}{0.32\linewidth}
		\centering
		\includegraphics[width=2.3in]{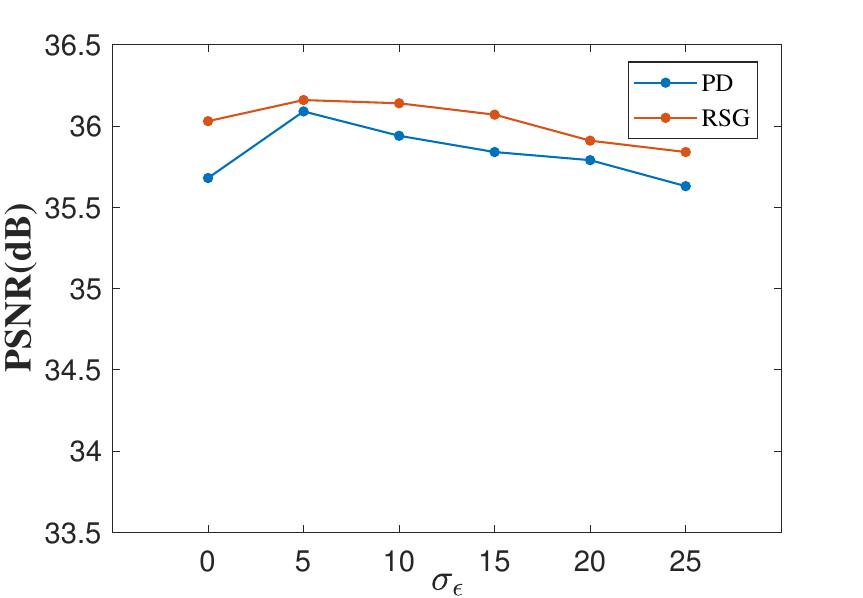}
		\vspace{-0.6cm}
		\caption{$L_{PBSN_{2}}$}
		\label{PBSN2}
	\end{subfigure}
	\hfill
	\begin{subfigure}{0.32\linewidth}
		\centering
		\includegraphics[width=2.3in]{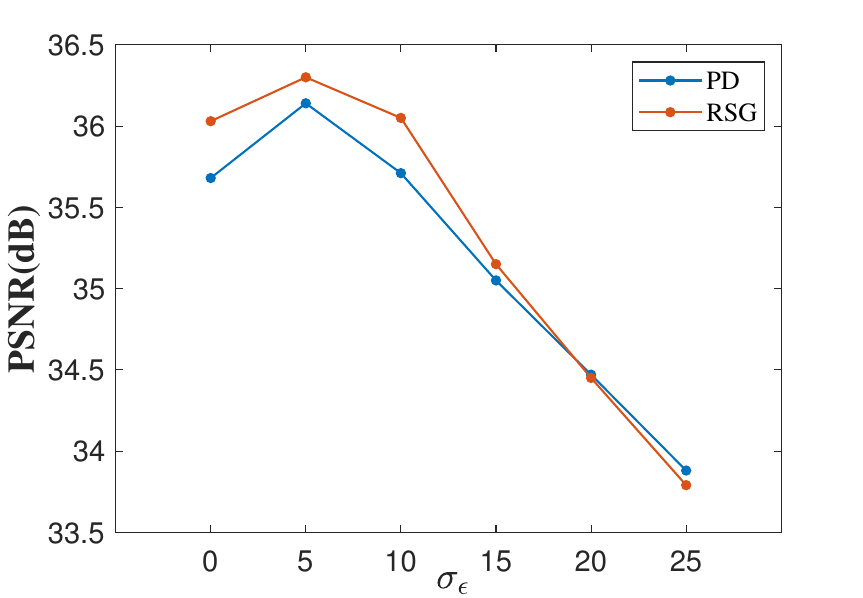}
		\vspace{-0.6cm}
		\caption{$L_{PBSN_{3}}$}
		\label{PBSN3}
	\end{subfigure}
	\vspace{-0.2cm}
	\caption{Ablation study of $\sigma_{\epsilon}$ in $L_{PBSN_{1,2,3}}$ for training on the SIDD validation dataset \cite{abdelhamed2018high}.} 
	\label{PBSN123}
\end{figure*}

\begin{figure}
	
	\centering
	\begin{subfigure}{0.15\linewidth}
		\centering
		\includegraphics[width=0.5in]{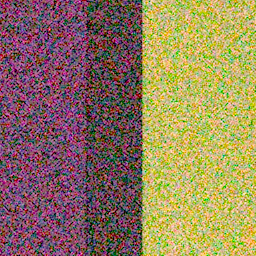}
		\vspace{-0.6cm}
		\caption{}
	\end{subfigure}
	\begin{subfigure}{0.15\linewidth}
		\centering
		\includegraphics[width=0.5in]{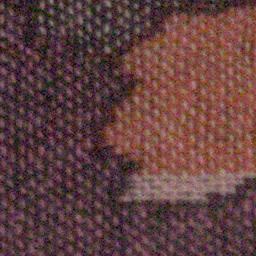}
		\vspace{-0.6cm}
		\caption{}
	\end{subfigure}
	\begin{subfigure}{0.15\linewidth}
		\centering
		\includegraphics[width=0.5in]{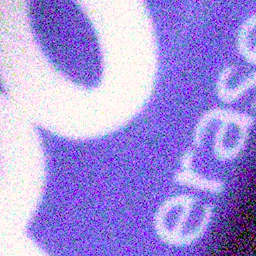}
		\vspace{-0.6cm}
		\caption{}
	\end{subfigure}
	\begin{subfigure}{0.15\linewidth}
		\centering
		\includegraphics[width=0.5in]{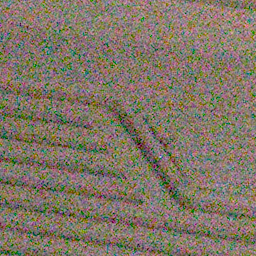}
		\vspace{-0.6cm}
		\caption{}
	\end{subfigure}
	\begin{subfigure}{0.15\linewidth}
		\centering
		\includegraphics[width=0.5in]{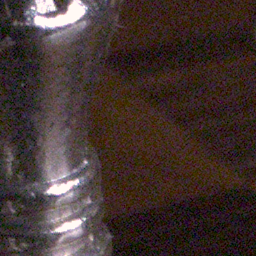}
		\vspace{-0.6cm}
		\caption{}
	\end{subfigure}
	\begin{subfigure}{0.15\linewidth}
		\centering
		\includegraphics[width=0.5in]{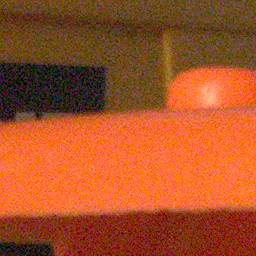}
		\vspace{-0.6cm}
		\caption{}
	\end{subfigure}
	
	\vspace{-0.1cm}
	\begin{subfigure}{\linewidth}
		\centering
		\includegraphics[width=3.2in]{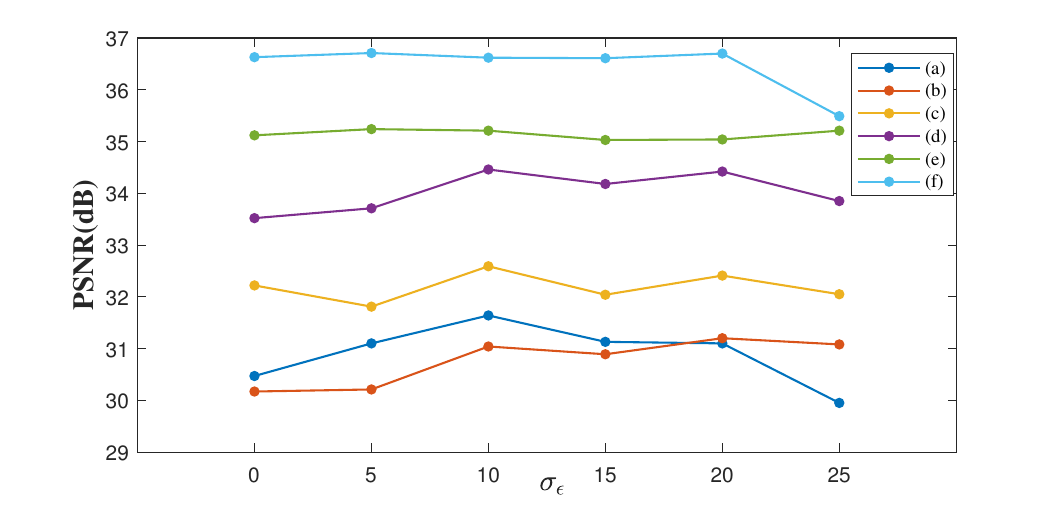}
		\label{}
	\end{subfigure}
	\vspace{-0.7cm}
	\caption{Ablation study of $\sigma_{\epsilon}$ in $L_{PBSN_{2}}$ for training in different images.} 
	\label{pd_img}
\end{figure}

\subsection{Perturbation in BSN}

In order to solve the above problems, we suggest adding perturbations to Eq. (\ref{APBSN}) to increase the number of training data and avoid fitting to the fixed pattern. We should ensure that the sub-images as inputs or targets remain consistent with the BSN assumption after adding perturbations. 
This is necessary for training the BSN using these sub-images for denoising purposes. A simple solution is using AWGN as the perturbation. Suppose $\epsilon$, $\epsilon_{1}$, and $\epsilon_{2}$ are all perturbations of a Gaussian distribution with mean 0 and variance $\sigma^{2}_{\epsilon}$. We propose three perturbation BSN losses $L_{PBSN_{1,2,3}}$ as follows:
\begin{equation}
	\begin{split}
		L_{PBSN_{1}}\! &= E_{y} \{\left \| B(P\!D(y) \!+\! \epsilon) - P\!D(y) \right \|_{1}\},
		\\
		L_{PBSN_{2}}\! &= E_{y} \{\left \| B(P\!D(y)) - (P\!D(y) \!+\! \epsilon) \right \|_{1}\},
		\\
		L_{PBSN_{3}}\! &= E_{y} \{\left \| B(P\!D(y) \!+\! \epsilon_{1}) - (P\!D(y) \!+\! \epsilon_{2}) \right \|_{1}\}.
	\end{split}
	\label{PBSN}
\end{equation}

We test the effect of different variances of $\epsilon$ or ($\epsilon_{1}, \epsilon_{2}$) in Eq. (\ref{PBSN}) on the BSN training, and the results are shown in Figures \ref{PBSN123} and \ref{pd_img}. We observe that adding suitable perturbations can achieve improvements in BSN performance. Therefore, each of the losses in Eq. (\ref{PBSN}) fits our vision and can obtain a better BSN by training. In Section \ref{4.3}, we analyze the influence of perturbations in detail. The results in Figure \ref{PBSN123} show that different perturbation levels' effects on BSN performance vary. The results in Figure \ref{pd_img} show that the appropriate optimal perturbation for different levels of noisy images also varies. Therefore, it is a challenge to choose the appropriate perturbation.

\subsection{Sampling Difference as Perturbation}

PD can produce a series of similar sub-images ($P\!D_{1}(y)$, $P\!D_{2}(y)$, ..., where $P\!D_{i}(\cdot)$ denotes the $i$-th sub-image obtained after PD). 
Since the PD sampling process does not overlap, the pixels in the sub-images that are at the same position are not located at the same position in the original image. As a result, there are certain differences between these sub-images, which we refer to as sampling difference.

Based on the previous analysis, it is evident that adding perturbation for training can enhance the performance of BSN, but how to add the appropriate perturbations is a challenge. Since there exists a sampling difference among the sub-images acquired through sampling, this difference can be viewed as a type of perturbation. 
First of all, according to \cite{lee2022ap}, the sub-images obtained through PD sampling are naturally in line with the BSN assumption. Next, the difference between $P\!D_{1}(y)$ and $P\!D_{2}(y)$ can be considered as the perturbation, so we do not need to add additional perturbations as in Eq. (\ref{PBSN}). Therefore, $P\!D_{1}(y)$ and $P\!D_{2}(y)$ can be used as the input and target in the BSN training, respectively. 
Finally, we propose a new sampling difference BSN loss $L_{SDBSN}$ as follows:
\begin{equation}
	\begin{split}
		L_{SDBSN}= E_{y} \{\left \| B(P\!D_{1}(y)) - P\!D_{2}(y) \right \|_{1}\}. \\
	\end{split}
	\label{ABSN}
\end{equation}

\begin{figure}
	\begin{subfigure}{0.205\linewidth}
		\centering
		\includegraphics[width=0.75in]{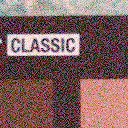}
	\end{subfigure}
	\begin{subfigure}{0.28\linewidth}
		\centering
		\includegraphics[width=0.75in]{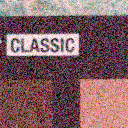}
	\end{subfigure}
	\begin{subfigure}{0.23\linewidth}
		\centering
		\includegraphics[width=0.75in]{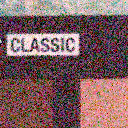}
	\end{subfigure}
	\begin{subfigure}{0.22\linewidth}
		\centering
		\includegraphics[width=0.75in]{image/RSG-1.png}
	\end{subfigure}

	\begin{subfigure}{0.49\linewidth}
		\centering		
		\includegraphics[width=0.75in]{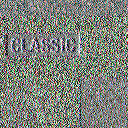}
		\vspace{-0.2cm}
		\caption{PD}
		\label{S-PD}
	\end{subfigure}
	\hfill
	\begin{subfigure}{0.49\linewidth}
	\centering
	\includegraphics[width=0.75in]{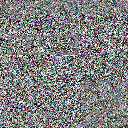}
	\vspace{-0.2cm}
	\caption{RSG}
	\label{S-RSG}
	\end{subfigure}
	\vspace{-0.3cm}
	\caption{Results of sampling by different methods. The second row is the sampling difference obtained by subtracting the two sub-samples in the first row.}
	\label{samples difference}
\end{figure}

\begin{figure}
	\centering
	\includegraphics[scale=0.4]{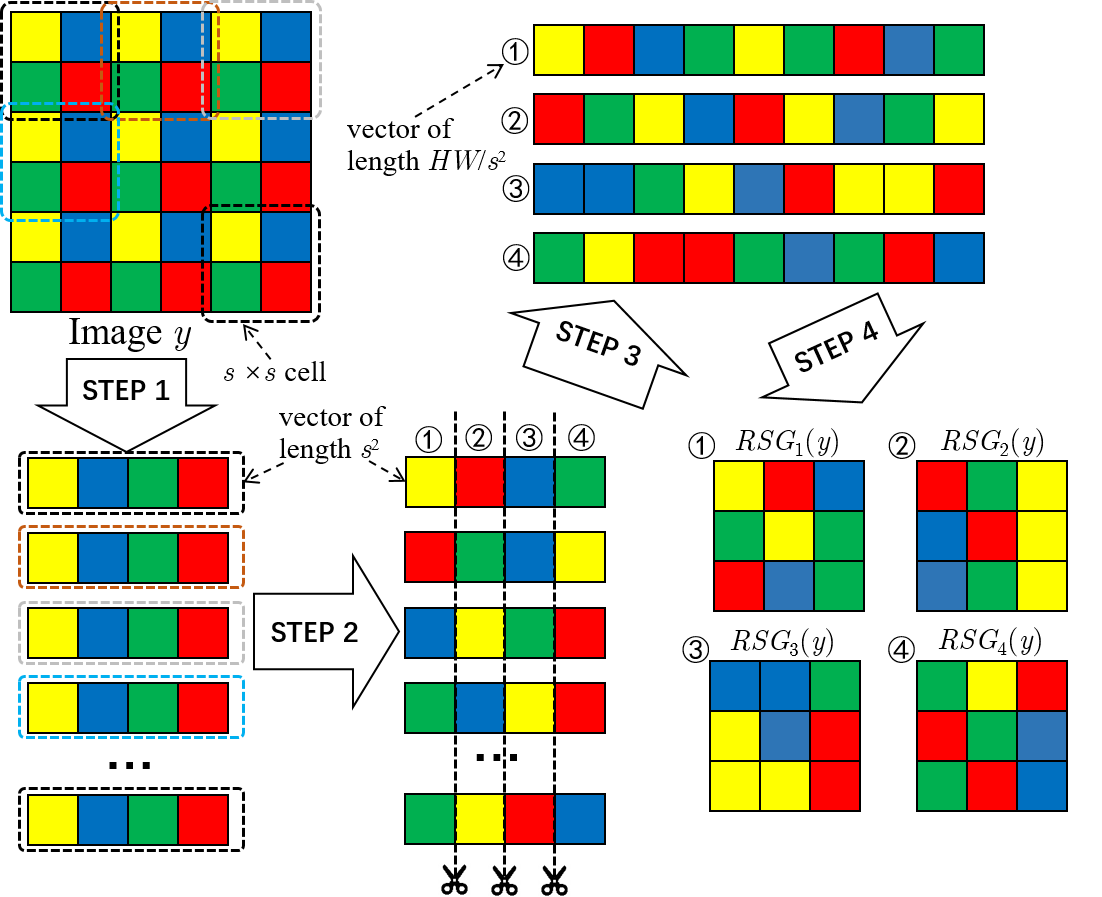}
	\vspace{-0.2cm}
	\caption{Example of generating sub-samples with RSG.} 
	\label{RSG}
\end{figure}

\begin{figure}
	\centering
	\begin{subfigure}{0.48\linewidth}
		\centering
		\includegraphics[width=0.75in]{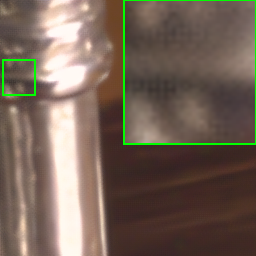}
		\includegraphics[width=0.75in]{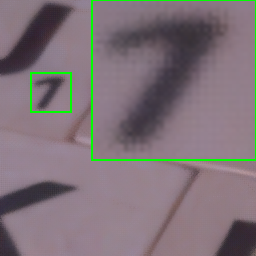}
		\caption{SDAP}
		\label{SDAP}
	\end{subfigure}
	\hfill
	\begin{subfigure}{0.48\linewidth}
		\centering
		\includegraphics[width=0.75in]{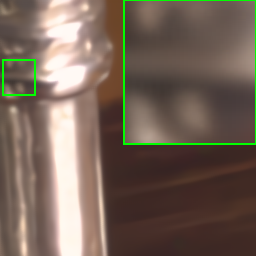}
		\includegraphics[width=0.75in]{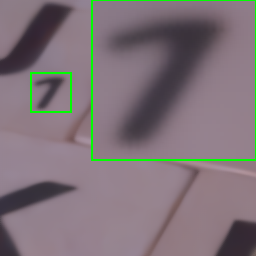}
		\caption{SDAP (E)}
		\label{SDAPE}
	\end{subfigure}
	\hfill
	\vspace{-0.3cm}
	\caption{Visual comparison between SDAP and SDAP (E).}
	\vspace{-0.2cm}
	\label{aliasing}
\end{figure}

\begin{figure*}
	\centering
	\includegraphics[scale=0.17]{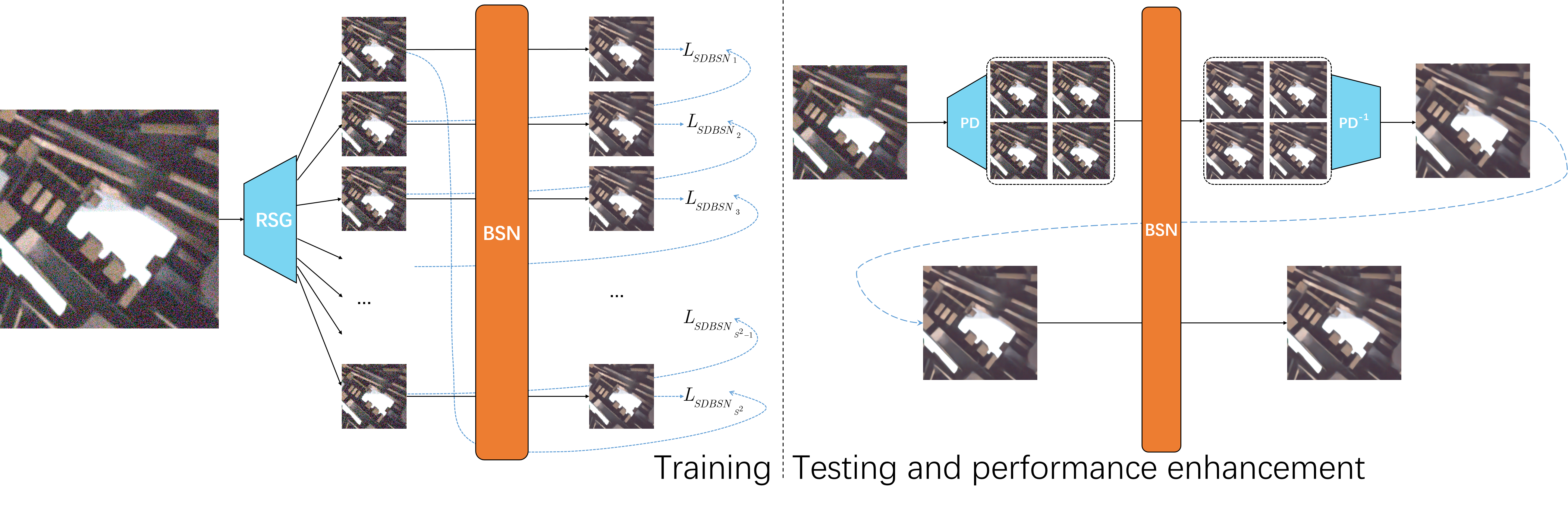}
	\vspace{-0.4cm}
	\caption{Overview of our proposed SDAP framework.} 
	\vspace{-0.3cm}
	\label{cheme}
\end{figure*}

\subsection{Random Sub-Samples Generation}

The sampling difference generated by the PD strategy can be obtained by subtracting the two sub-images generated by the PD. The sub-images and the sampling difference generated by PD are shown in Figure \ref{S-PD}. It can be observed that the sampling difference generated by PD is similar to the gradient map. Meanwhile, the PD strategy with a stride factor of $s$ can only generate  $s^2$ fixed sub-images for the same image, which still faces the data-hungry issue for BSN training. To obtain more sub-samples and make the sampling difference closer to random perturbation, we propose the random sub-samples generation strategy (RSG). 

The diagram of sub-images generation with the RSG is shown in Figure \ref{RSG}. Assume that the noisy image ($y$) length and width are $H$ and $W$, respectively. The details of RSG are described below:
\begin{enumerate}
	\item \textbf{Divide.} The image $y$ is divided into $HW/s^{2}$ non-overlapping cells with size $s \times s$. Then, we stretch the cells into vectors of length $s^{2}$. For better visualization, we set $s = 2$ in Figure \ref{RSG}. 
	\item \textbf{Shuffle.} Randomly shuffle the elements within each vector. Since this process is strictly random, the sub-samples obtained are different each time.
	\item \textbf{Reform.} Take the elements at the same position of each vector to form $s^{2}$ new vectors of length $HW/s^{2}$.
	\item \textbf{Reshape.} Reshape each vector into a sub-sample of size  $[W/s] \times [H/s]$. In this way, $s^{2}$  similar sub-samples ($R\!S\!G_{1}(y)$, $R\!S\!G_{2}(y)$, ... , $R\!S\!G_{s^{2}}(y)$) are obtained.
\end{enumerate}

To maximize the variance of each sub-sample, our sampling is non-overlapping.

 The sub-samples generated by the RSG strategy and their sampling difference is shown in Figure \ref{S-RSG}. It can be observed that the use of RSG does make the sampling difference more random. 
 Furthermore, we provide detailed analysis for this phenomenon. Both PD and RSG sample the image directly, so the calculation of the sampling difference can be considered as the deviation in the direction of the line connecting the sampling pixels of the two sub-samples. Since each pixel of the PD sub-sample has a fixed position in the original image, the difference has a fixed offset direction. Due to random sampling, the sampling difference of RSG sub-samples are not fixed in the direction of each pixel deviation, making the difference more random. In addition, as can also be seen from Figure \ref{RSG}, PD is equivalent to obtaining four fixed sub-images of yellow, blue, green and red. While RSG obtains random sub-images of \textcircled{{\small 1}}\textcircled{\small 2}\textcircled{\small 3}\textcircled{\small 4} with different results for each sampling, which can provide a large number of samples for training. Therefore, RSG strategy can also be considered as a solution to the data-hungry issue of training BSN with real noisy images.

\subsection{Cyclic Sampling Difference BSN Loss}

According to Eq. (\ref{ABSN}), only two sub-samples are needed. However, because multiple sub-samples are generated by RSG, using the whole sub-images instead of only two samples in training can better utilize the features of the generated sub-samples, leading to higher performance. Moreover, if only two sub-samples are used for training, the loss of each training iteration will fluctuate seriously, and the training of BSN cannot converge. To preserve the stability of training and to make full use of each sample, we introduce the strategy of cyclic and RSG to propose a new cyclic samples difference BSN loss $L_{CSDBSN}$:
\begin{equation}
	\begin{split}
		L&_{CSDBSN}	=  \sum_{i=1}^{s^{2}}\underset{cross-pairing}{\underbrace{\left \| B(R\!S\!G_{i}(y)) - R\!S\!G_{i+1}(y) \right \|_{1}}}\\
		& \ \ \ \ \ \ \ \ \ \ \ \  (R\!S\!G_{s^{2}+1}(y) = R\!S\!G_{1}(y)).
	\end{split}
	\label{CSDBSN}
\end{equation}

Since the sub-samples are generated randomly, the training data is also random. The target corresponding to each input in $L_{CSDBSN}$ is different. Therefore, two sub-samples are avoided to form a fixed mapping during the cycle. This cyclic loss has the following merits: 1) it imposes constraints on the full pixel of the original noisy image; 2) it ensures that all the sub-samples generated by RSG are well exploited; 3) it makes the training of BSN more robust. Figure \ref{cheme} visually illustrates our training scheme with the cyclic sampling difference BSN loss.

\subsection{Proposed SDAP Framework}
For the training stage, we first sample the noisy images into $s^{2}$ sub-samples by RSG. Then, we denoise sub-samples by BSN. The loss is calculated by cross-pairing the sub-samples after denoising with those before denoising. Finally, the above steps are iterated, and the loss function is updated to optimize the BSN until it converges. The training process is shown in the left half of Figure \ref{cheme}.

For the testing stage, due to the randomness of the single RSG, the test results are not fixed. Since the sampling pattern of PD is fixed, the test results are stable for testing. Therefore, for default setting, we directly adopt the strategy of PD. We denoise the sub-images after PD and then stitch them to obtain the final denoised image.

The PD strategy is introduced in the testing stage to break the spatial correlation of real-world noise. However, this also makes the pixels of the denoised image discontinuous and produces checkerboard artifacts as shown in Figure \ref{SDAP}.  To make the denoised image have better visual performance, we propose to denoise the initial denoised image again by BSN to remove the unpleasing artifacts. This performance enhancement method is denoted by ``SDAP (E)" and is shown in the right half of Figure \ref{cheme}. The comparison of the final visual effect is shown in Figure \ref{SDAPE}. The results show that our proposed test strategy can remove checkerboard artifacts and make the images more natural. More details about PD and RSG for testing are provided in \textbf{Supplementary Material}.

\begin{table*} \small
	
	\vspace{-0.10cm}
	\centering
	\setlength{\tabcolsep}{2mm}
	\begin{tabular}{cclccc}
		\toprule
		Type of supervision & Training data &\multirow{1}{*}{Method}    & SIDD validation & SIDD benchmark  & DND benchmark  \\
		\midrule
		\multirow{2}{*}{Non-learning based} &\multirow{2}{*}{-}&BM3D \cite{dabov2007image}& 31.75/0.7061 & 25.65/0.685  & 34.51/0.8507\\
		&& WNNM \cite{gu2014weighted} & 26.31/0.5240 & 25.78/0.809 & 34.67/0.8646\\
		\midrule
		\multirow{7}{*}{Supervised} &\multirow{7}{*}{Paired noisy-clean}&DnCNN \cite{zhang2017beyond}& 26.20/0.4414 & 23.66/0.583 & 32.43/0.7900 \\
		&&TNRD \cite{chen2016trainable}& 26.99/0.7440 & 24.73/0.643 & 33.65/0.8306\\
		&&CBDNet \cite{guo2019toward}& 30.83/0.7541 & 33.28/0.868 & 38.06/0.9421\\
		&&RIDNet \cite{anwar2019real} & 38.76/0.9132 & 37.87/0.943 & 39.25/0.9528\\
		&&VDN \cite{yue2019variational}& 39.29/0.9109 & 39.26/0.955  & 39.38/0.9518\\
		&&Zhou \etal~\cite{zhou2020awgn} & - & 34.00/0.898 & 38.40/0.945 \\
		&&DeamNet \cite{ren2021adaptive}& 39.40/0.9169 & 39.35/0.955 & 39.63/0.9531\\
		\midrule
		\multirow{5}{*}{Pseudo-supervised} &\multirow{3}{*}{Unpaired noisy-clean}&GCBD \cite{chen2018image}& - & -  & 35.58/	0.9217\\
		&&D-BSN \cite{wu2020unpaired}& - & - & 37.93/0.9373\\
		&&C2N \cite{jang2021c2n}& 35.36/0.8901 & 35.35/0.937 & 37.28/0.9237\\
		\cmidrule{2-6}
		& Paired noisy-noisy & R2R \cite{pang2021recorrupted}& 35.04/0.8440 & 34.78/0.898 & 37.61/0.9368\\
		\midrule
		\multirow{9}{*}{Self-supervised}& \multirow{9}{*}{Single noisy} &N2V \cite{krull2019noise2void}& 29.35/0.6510 & 27.68/0.668 & -\\
		&&N2S \cite{batson2019noise2self}& 30.72/0.7870 & 29.56/0.808 & -\\
		&&NAC \cite{xu2020noisy}& - & - & 36.20/0.9252\\
		&&Neighbor2Neighbor \cite{huang2021neighbor2neighbor} & 28.00/0.5890 & 27.96/0.670 & 31.40/0.7880\\
		&&CVF-SID \cite{neshatavar2022cvf}& 34.17/0.8719 & 34.71/0.917 & 36.50/0.9233\\
		&&AP-BSN \cite{lee2022ap}& 34.46/0.8501 & 34.90/0.900  & 37.46/0.9244\\
		&&\textbf{SDAP (Ours)} & \textbf{36.58/0.8630}	&	\textbf{36.54/0.919}	&	\textbf{37.71/0.9278} \\
		&&\textbf{SDAP (S) (Ours)} & \textbf{36.71/0.8640}	&	\textbf{36.68/0.919}	&	\textbf{38.18/0.9322} \\
		&&\textbf{SDAP (E) (Ours}) & \textbf{37.30/0.8937} & \textbf{37.24/0.936} & \textbf{37.86/0.9366} \\ 
		&&\textbf{SDAP (S)(E) (Ours}) & \textbf{37.55/0.8943} & \textbf{37.53/0.936} & \textbf{38.56/0.9402} \\
		\bottomrule
		\vspace{-0.4cm}
	\end{tabular}
	\caption{Quantitative PSNR(dB)/SSIM Results on SIDD and DND Dataset.}
	\label{test}
\end{table*}

\section{Experiments}
\subsection{Implementation Details}
We use the general BSN architecture \cite{wu2020unpaired,lee2022ap} in our method. According to \cite{lee2022ap}, we empirically set the stride factor $s$ of RSG to 5 for training. To ensure the stability of the test, we still use the PD with stride factor 2 for testing and ablation study. During the training, we use Adam as the optimizer with default settings. The initial learning rate is 0.0001, batch size and patch size are set to 16 and 160×160 respectively, and epochs are taken as 15. 25600 patches are used in one epoch. To fine-tune the network, BSN continues to be optimised for 10 epochs. Meanwhile, we decay the learning rate by a factor of 10, reduce batch size to 8, and adjust patch size to 250×250. We use PyTorch to implement our BSN and train it on an NVIDIA RTX 3090 GPU.

\subsection{Dataset}
\paragraph{SIDD \cite{abdelhamed2018high}.} The SIDD dataset uses five representative mobile phone cameras to capture approximately 300,000 noisy images for 10 scenes under different lighting conditions and obtain the corresponding relatively clean images by specific statistical methods. The SIDD dataset provides an SIDD Medium dataset of 320 data pairs for training, an SIDD validation dataset of 1280 image blocks of size 256×256 for validation, and an SIDD benchmark dataset of 40 high-resolution noisy images for testing.

\paragraph{DND \cite{plotz2017benchmarking}.} The DND dataset contains 50 high-resolution test image pairs taken by consumer-grade cameras of various sensor sizes. However, because the images are too large, DND cropped all the images to 512×512 image blocks, giving 1000 image blocks for testing. The DND dataset does not provide training images, and the clean images of the test dataset are not publicly available.

Note that unless otherwise stated, we used SIDD Medium dataset for training in the following experiments.

\subsection{Analyzing Perturbation}
\label{4.3}

We first validate the effect of the perturbation in $L_{AP\!-\!BSN}$ for real-world sRGB image denoising. We train the BSN by $L_{PBSN_{1,2,3}}$ with different variances of $\epsilon$, i.e., $\sigma_{\epsilon} \in \{0, 5, 10, 15, 20, 25\}$. Meanwhile, we also replace PD with RSG for a similar set of experiments. Figure \ref{PBSN123} shows the PSNR(dB) results in the SIDD validation set. 

We observe that BSN performs best on SIDD validation dataset when $\sigma_{\epsilon} = 5$ for both PD and RSG. After this, the performance decreases as $\sigma_{\epsilon}$ increases. Specifically, for $L_{PBSN_{1,2,3}}$, since the perturbation is added to the training data, when the perturbation is too large, it will significantly change the noise distribution of the training image, making it difficult for the BSN to adapt to the original noise. Therefore, the denoising performance of BSN decreases significantly when $\sigma_{\epsilon} > 15$. Also, since RSG and perturbation have similar roles, both are designed to obtain more training data. When the perturbation is too large, the role of RSG is difficult to be reflected. Therefore, when the perturbation is small, the RSG strategy outperforms the PD strategy. When the perturbation is large, the RSG strategy and the PD strategy are used for BSN to obtain similar denoising results.

In addition, we note that the most suitable perturbation standard deviation $\sigma_{\epsilon}$ is different for each of these images in Figure \ref{pd_img}. Nevertheless, when we consider the sampling difference as the perturbation, the BSN performs the best regardless of the data. In Section \ref{4.5}, we analyze the sampling difference as perturbation in detail. Therefore, Eq. (\ref{CSDBSN}) is the final loss function of our proposed method.

\begin{figure*}
	
	\centering
	\begin{subfigure}{0.12\linewidth}
		\centering
		\includegraphics[width=0.8in]{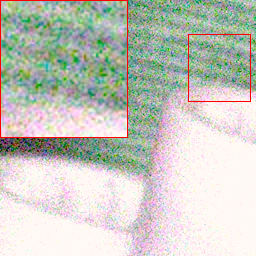}
		\includegraphics[width=0.8in]{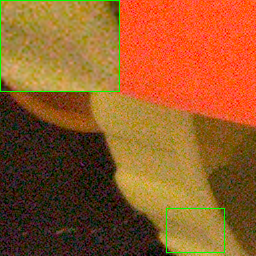}
		\includegraphics[width=0.8in]{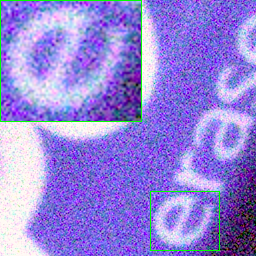}
		\scriptsize{Noisy}
	\end{subfigure}
	\begin{subfigure}{0.12\linewidth}
		\centering
		\includegraphics[width=0.8in]{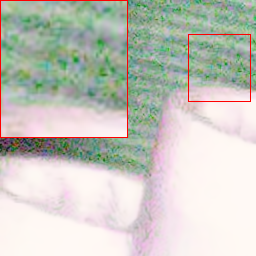}
		\includegraphics[width=0.8in]{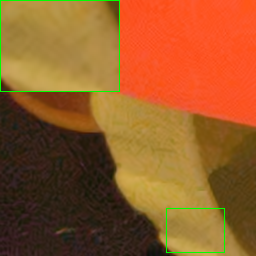}
		\includegraphics[width=0.8in]{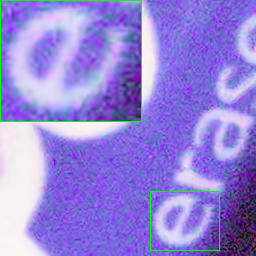}
		\scriptsize{BM3D \cite{dabov2007image}}
	\end{subfigure}
	\begin{subfigure}{0.12\linewidth}
		\centering
		\includegraphics[width=0.8in]{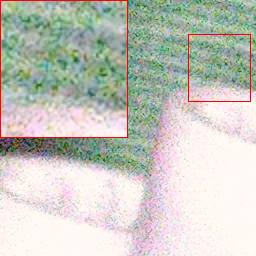}
		\includegraphics[width=0.8in]{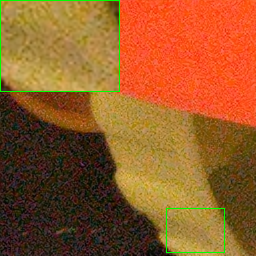}
		\includegraphics[width=0.8in]{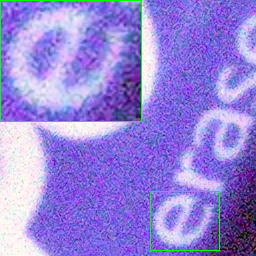}
		\scriptsize{DnCNN \cite{zhang2017beyond}}
	\end{subfigure}
	\begin{subfigure}{0.12\linewidth}
		\centering
		\includegraphics[width=0.8in]{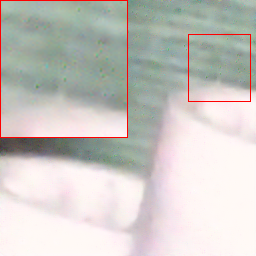}
		\includegraphics[width=0.8in]{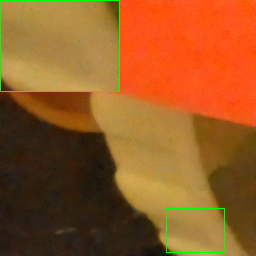}
		\includegraphics[width=0.8in]{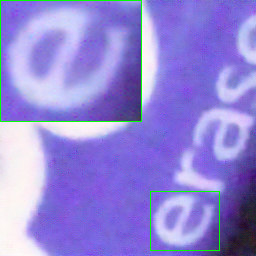}
		\scriptsize{CBDNet \cite{guo2019toward}}
	\end{subfigure}
	\begin{subfigure}{0.12\linewidth}
		\centering
		\includegraphics[width=0.8in]{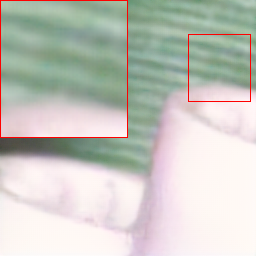}
		\includegraphics[width=0.8in]{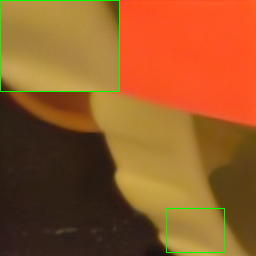}
		\includegraphics[width=0.8in]{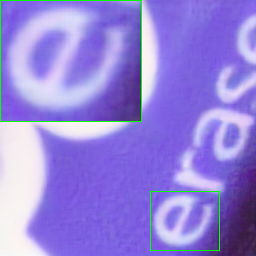}
		\scriptsize{C2N \cite{jang2021c2n}}
	\end{subfigure}
	\begin{subfigure}{0.12\linewidth}
		\centering
		\includegraphics[width=0.8in]{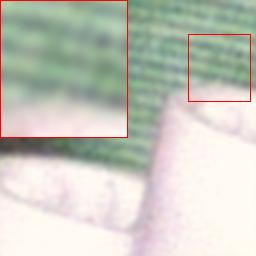}
		\includegraphics[width=0.8in]{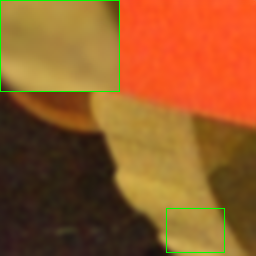}
		\includegraphics[width=0.8in]{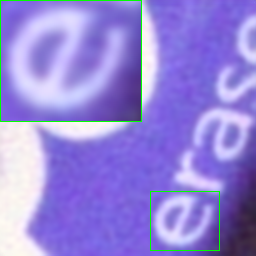}
		\scriptsize{CVF-SID \cite{neshatavar2022cvf}}
	\end{subfigure}
	\begin{subfigure}{0.12\linewidth}
		\centering
		\includegraphics[width=0.8in]{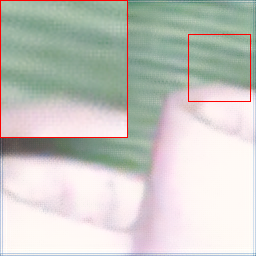}
		\includegraphics[width=0.8in]{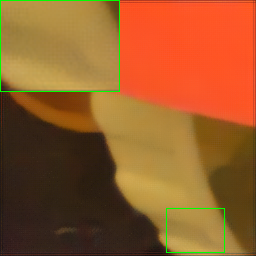}
		\includegraphics[width=0.8in]{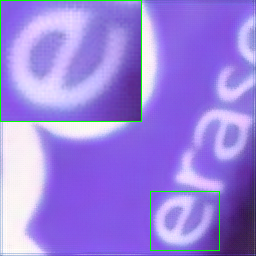}
		\scriptsize{AP-BSN \cite{lee2022ap}}
	\end{subfigure}
	\begin{subfigure}{0.12\linewidth}
		\centering
		\includegraphics[width=0.8in]{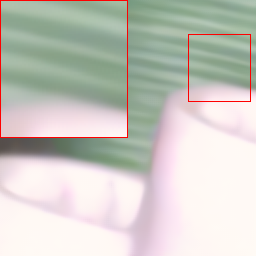}
		\includegraphics[width=0.8in]{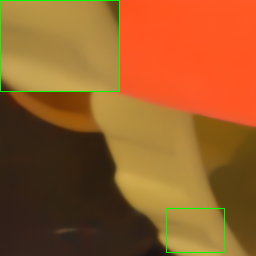}
		\includegraphics[width=0.8in]{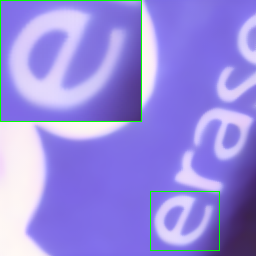}
		\scriptsize{\textbf{SDAP (S)(E) (Ours)}} 
	\end{subfigure}
	\vspace{-0.1cm}
	\caption{Visual comparison of our method against other competing methods on the SIDD validation dataset \cite{abdelhamed2018high}.} 
	\label{sidd}
\end{figure*}

\subsection{Denoising of Real Image}

We use SIDD validation, SIDD benchmark, and DND benchmark datasets to evaluate the performance of our SDAP in real-world denoising tasks. For the SIDD validation dataset, because the paired noisy-clean images are provided, we can directly test the PSNR/SSIM results. For the SIDD and DND benchmarks, we submit the denoised images to websites for server-side evaluation to obtain the final results (PSNR/SSIM values). To evaluate our model on the three different datasets, we adopt two different strategies for training.

\paragraph{Training on SIDD Medium dataset.} 
We train our self-supervised method with noisy images from the SIDD Medium dataset. Then, the obtained model is evaluated on different datasets. This training method is represented as ``SDAP" in Table \ref{test}. 

\paragraph{Training on test dataset.} 
Since our method is fully self-supervised \cite{neshatavar2022cvf}, instead of using the fixed SIDD Medium dataset for training, we can use the test images for self-supervised training as well. Therefore, we train our method on three test datasets in self-supervised manner, respectively. The trained models are evaluated with the corresponding test datasets. Since the noise distribution of training and test images are identical, this strategy can obtain a BSN model with better performance. Fully self-supervised training method is represented as ``SDAP (S)" in Table \ref{test}.

It should be noted that ``(S)" denotes the method to improve the performance in training, while ``(E)" denotes the way to improve the performance in testing. Thus ``SDAP (S)(E)" is the way in which the proposed method can achieve the best performance.

The results in Table \ref{test} show that this method achieves significantly better results than some traditional methods, such as BM3D, on the SIDD Benchmark test data. Even when compared with some supervised deep learning-based image denoising methods, this method achieves superior denoising results. For example, the PSNR and SSIM results of this method are 3.36dB and 0.052 higher compared with CBDNet. When we compared with the SOTA self-supervised image denoising approaches, our method also has obvious advantages. For example, compared with AP-BSN, which has the same network structure as us, SDAP achieves 1.74dB and 0.020 increases in PSNR and SSIM metrics, respectively. The performance of SDAP on the DND dataset has similar findings as those on SIDD. Figures \ref{dnd} and \ref{sidd} provides visual comparisons of several methods, and the results further validate the visual superiority of our method to other methods.

\begin{table}\small
	\centering
	\setlength{\tabcolsep}{2mm}
	\begin{tabular}{ccccc}
		\toprule
		PD & RSG  & $L_{AP\!-\!BSN}$ & $L_{CSDBSN}$ & PSNR(dB)/SSIM\\
		\midrule
		\checkmark & - & \checkmark & - & 35.68/0.8382 \\
		\checkmark & - & - & \checkmark & 36.19/0.8472 \\
		- & \checkmark & \checkmark & - & 36.03/0.8533 \\
		\midrule
		- & \checkmark & - & \checkmark & \textbf{36.58/0.8630} \\
		\bottomrule
		\vspace{-0.41cm}
	\end{tabular}
	\caption{Investigation of RSG and $L_{CSDBSN}$ on SIDD Validation Dataset. }
	\label{ablation}
\end{table}

\subsection{Ablation Study}
\label{4.5}
We examine two major determinants of our model: a) RSG; b) $L_{CSDBSN}$. If there is a ``\checkmark" in the column of $L_{CSDBSN}$ (or $L_{APBSN}$), it means that $L_{CSDBSN}$ (or $L_{APBSN}$) is used for training, and vice versa. If there is a ``\checkmark" in the column of RSG, it means that the RSG strategy is used for training, and we replace all $P\!D$ in the loss with $R\!S\!G$. If there is a ``\checkmark" in the column of PD, it means that the PD strategy is used for training, and we replace all $R\!S\!G$ in the loss with $P\!D$. 

\paragraph{Study of RSG.}
The performance of training with RSG in Table \ref{ablation} is always higher than training with PD. 
A comparison between the first and third rows of Table \ref{ablation} shows that replacing PD with RSG increases the PSNR/SSIM result by 0.35dB/0.0151. The comparison between the second and fourth rows shows a similar conclusion. These results verify the effectiveness for RSG. 

\paragraph{Study of $L_{CSDBSN}$.}
Rows 2 and 4 of Table \ref{ablation} show the effect of our proposed $L_{CSDBSN}$, where our proposed loss improves the denoising performance by 0.51dB/0.0090 in the case of training with PD and 0.55dB/0.0097 in the case of training with RSG. It is worth noting that, irrespective of whether PD or RSG is used in $L_{CSDBSN}$ for BSN training, the performance of BSN is superior to that achieved by directly adding Gaussian perturbations in $L_{AP\!-\!BSN}$ (i.e., using $L_{CSDBSN}$ for training is better than using $L_{PBSN_{1,2,3}}$).

\section{Conclusion}
In this paper, we first analyze the reasons for the limited performance of BSN when used for real image denoising. Based on this, we propose to add perturbations to the training data and consider sampling difference as perturbation. Further, we propose SDAP framework with random sub-samples generation and cyclic sampling difference loss. Our SDAP does not require clean images for training and outperforms existing pseudo-supervised/self-supervised methods. We believe that our approach can provide promising inspirations for various self-supervised real-world denoising methods.

\paragraph{Acknowledgement.} This work was supported by the National Natural Science Foundation of China under Grant 62171304.

{\small
\bibliographystyle{ieee_fullname}
\bibliography{egbib}
}

\balance

\newpage

\section*{S1. RSG for Testing}

\begin{figure}
	\centering
	\includegraphics[scale=0.146]{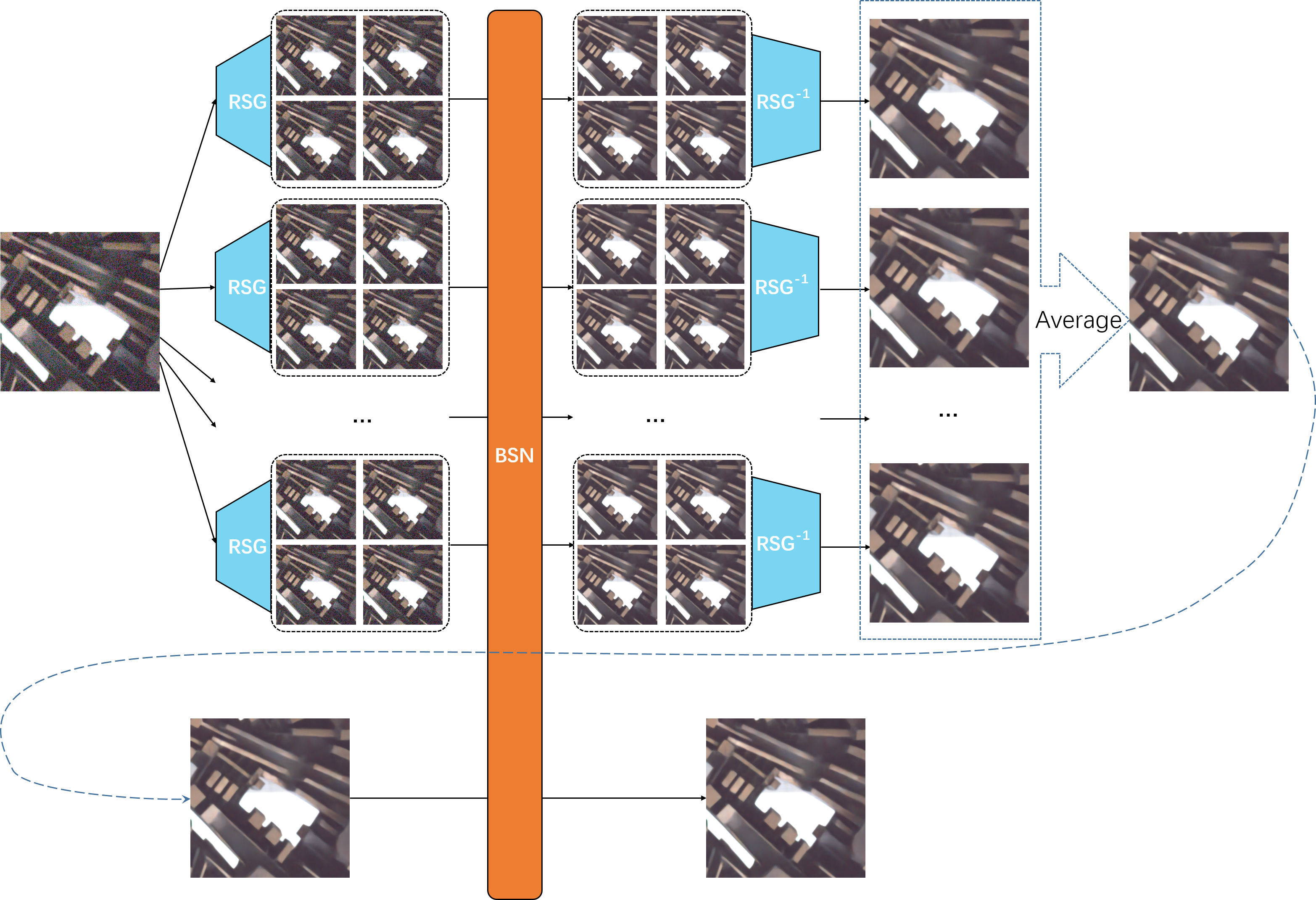}
	\caption{Overview of ``n RSG + enhancement" for testing. RSG$^{-1}$ is the inverse operator of RSG.} 
	\label{nrsg}
\end{figure}

\begin{figure}
	\centering
	\begin{subfigure}{0.45\linewidth}
		\centering
		\includegraphics[width=1.5in]{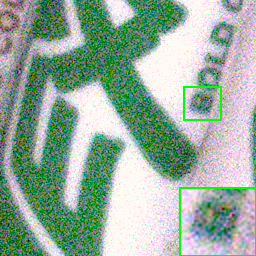}
		\includegraphics[width=1.5in]{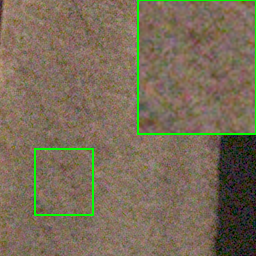}
		\caption{Noisy}
	\end{subfigure}
	\hfill
	\begin{subfigure}{0.45\linewidth}
		\centering
		\includegraphics[width=1.5in]{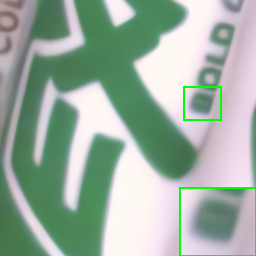}
		\includegraphics[width=1.5in]{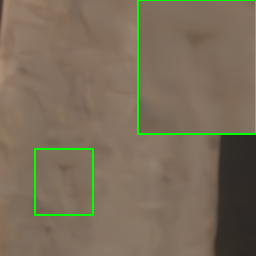}
		\caption{SDAP (S)(E)}
	\end{subfigure}
	\vspace{-0.21cm}
	\caption{The limitations of our method. It can be seen that the denoising results of our method are relatively smooth to some extent.}
	\label{Limitations}
\end{figure}

To evaluate the impact of our RSG strategy on the test, we replace PD with RSG in the testing stage. Table \ref{pd/rsg} shows the PSNR and SSIM results in the SIDD validation dataset \cite{abdelhamed2018high} for different testing methods. 

\begin{table}
	\centering
	\setlength{\tabcolsep}{3.5mm}
	\begin{tabular}{ccc}
		\toprule
		Testing strategy & RSG & PD\\
		\midrule
		SDAP & 36.19/0.8585 &  \textbf{36.58/0.8630}\\
		SDAP (E) & 37.22/0.8926 &  \textbf{37.30/0.8937}\\
		\bottomrule
	\end{tabular}
	\caption{The effect of different testing strategies on PSNR(dB)/SSIM. }
	\label{pd/rsg}
\end{table}

\begin{table} \footnotesize
	\centering
	\setlength{\tabcolsep}{0.38mm}
	\begin{tabular}{ccccccccc}
		\toprule
		n  & 1 & 2 & 3 & 4 & 5 & 6 & 7 & 8\\
		\midrule
		n RSG& 36.19 & 36.52 & 36.64 & 36.70 & 36.74 & 36.76 & 36.78 & 36.79\\
		n RSG + enhancement& 37.22 & 37.24 & 37.25 & 37.25 & 37.25 & 37.25 & 37.25 & 37.25\\
		\bottomrule
	\end{tabular}
	\caption{The effect of different ``n" values on PSNR(dB).}
	\label{n test}
\end{table}

In training, RSG generates a larger quantity of training data with more varied random sampling differences as compared to PD. Thus, RSG can achieve better results when used for training. However, in testing, while sampling helps to break the spatial correlation of noise pixels, it also reduces the spatial correlation of signal pixels to some degree. As a result of its random sampling step, the RSG strategy tends to have a more negative impact on the spatial correlation of signal pixels, which can result in a decline in BSN denoising performance. Consequently, the performance of testing with RSG is inferior to that of testing with PD.

Since the sub-samples generated by RSG are different each time, the strategy of averaging after multiple denoising can be used to achieve better performance of the final denoising results. Therefore, we use RSG n times for testing and average the results before enhancement. This performance enhancement method is denoted by ``n RSG + enhancement", which is shown schematically in Figure \ref{nrsg}. ``n RSG" means that no enhancement is performed and the denoising results are averaged directly. Table \ref{n test} shows the effect of different ``n" values on the PSNR results when RSG is used for testing.

The results shown in Table \ref{n test} verify that the using ``single PD + augmentation" (SDAP (S)(E)) in testing is always better than RSG. Therefore, we use PD strategy instead of RSG strategy in testing.

\section*{S2. More Results on Real-world Datasets}

In Figures \ref{sidd} and \ref{dnd}, we show more denoising results on the SIDD validation dataset \cite{abdelhamed2018high} and DND benchmark \cite{plotz2017benchmarking}.

\begin{figure*} \scriptsize
	
	\centering
	\begin{subfigure}{0.12\linewidth}
		\centering
		\includegraphics[width=0.8in]{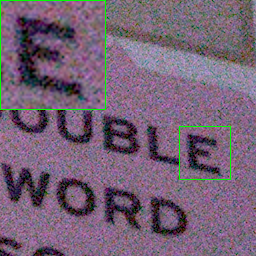}
		PSNR/SSIM
		\includegraphics[width=0.8in]{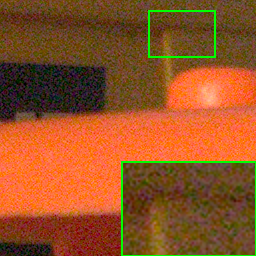}
		PSNR/SSIM
		\includegraphics[width=0.8in]{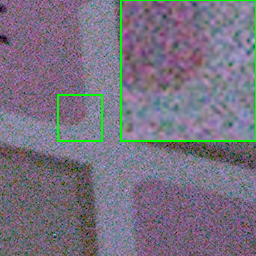}
		PSNR/SSIM
		\includegraphics[width=0.8in]{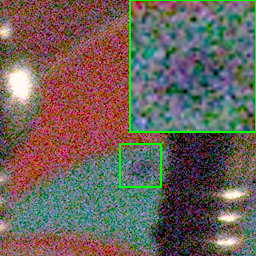}
		PSNR/SSIM
		\includegraphics[width=0.8in]{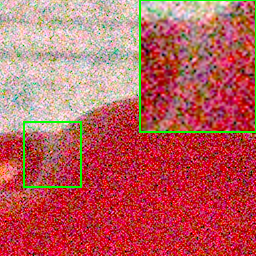}
		PSNR/SSIM
		\includegraphics[width=0.8in]{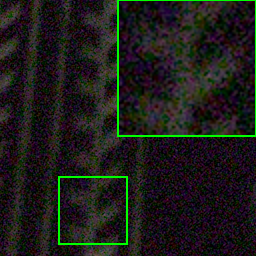}
		PSNR/SSIM
		
		\scriptsize{Noisy}
	\end{subfigure}
	\begin{subfigure}{0.12\linewidth}
		\centering
		\includegraphics[width=0.8in]{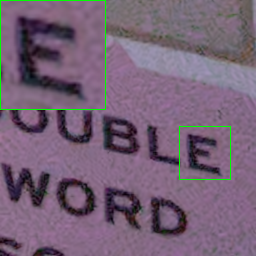}
		30.10dB/0.7877
		\includegraphics[width=0.8in]{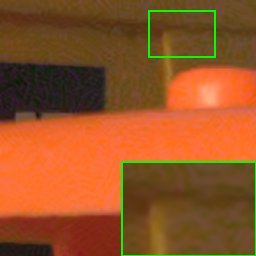}
		32.42dB/0.8524
		\includegraphics[width=0.8in]{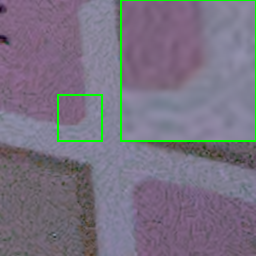}
		30.95dB/0.7308
		\includegraphics[width=0.8in]{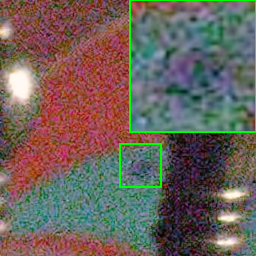}
		21.52dB/0.2929
		\includegraphics[width=0.8in]{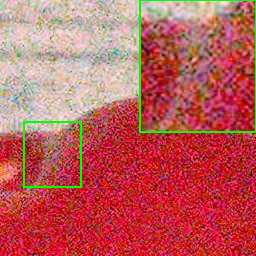}
		20.32dB/0.2544
		\includegraphics[width=0.8in]{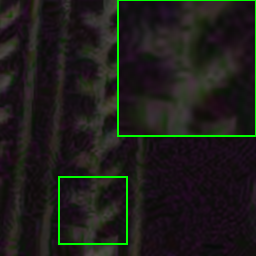}
		32.29/0.8361
		
		\scriptsize{BM3D \cite{dabov2007image}}
	\end{subfigure}
	\begin{subfigure}{0.12\linewidth}
		\centering
		\includegraphics[width=0.8in]{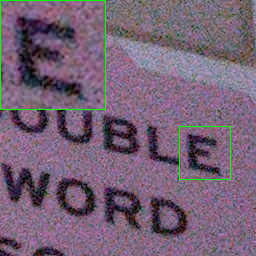}
		24.44dB/0.4195
		\includegraphics[width=0.8in]{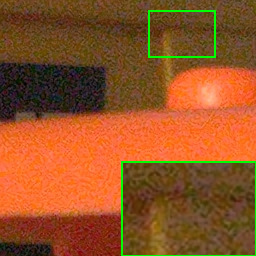}
		27.13dB/0.5097
		\includegraphics[width=0.8in]{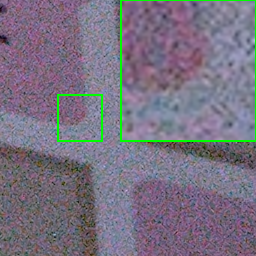}
		24.44dB/0.3156
		\includegraphics[width=0.8in]{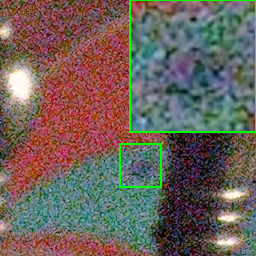}
		20.06dB/0.2079
		\includegraphics[width=0.8in]{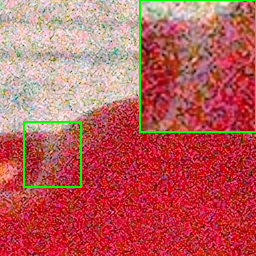}
		19.39dB/0.1865
		\includegraphics[width=0.8in]{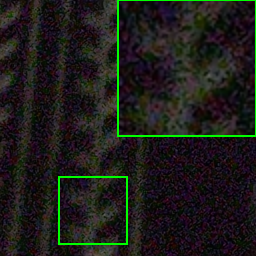}
		26.66/0.4318
		
		\scriptsize{DnCNN \cite{zhang2017beyond}}
	\end{subfigure}
	\begin{subfigure}{0.12\linewidth}
		\centering
		\includegraphics[width=0.8in]{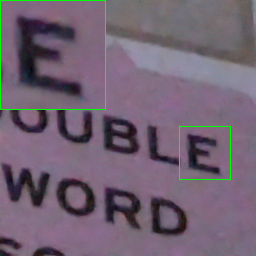}
		30.17dB/0.8632
		\includegraphics[width=0.8in]{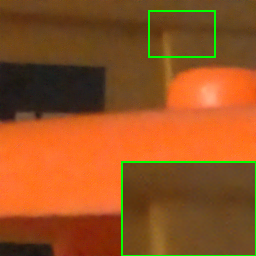}
		29.61dB/0.8038
		\includegraphics[width=0.8in]{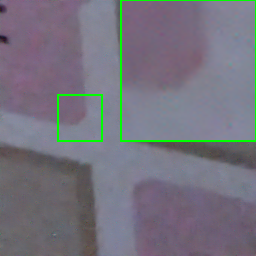}
		31.73dB/0.8362
		\includegraphics[width=0.8in]{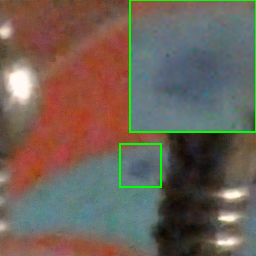}
		24.40dB/0.6136
		\includegraphics[width=0.8in]{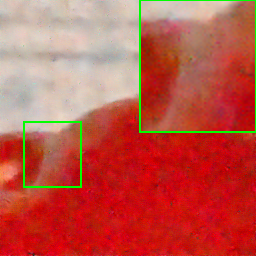}
		21.35dB/0.4500
		\includegraphics[width=0.8in]{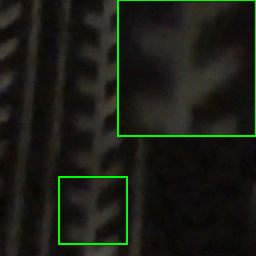}
		29.51/0.8150
		
		\scriptsize{CBDNet \cite{guo2019toward}}
	\end{subfigure}
	\begin{subfigure}{0.12\linewidth}
		\centering
		\includegraphics[width=0.8in]{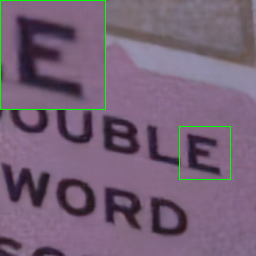}
		34.06dB/0.9145
		\includegraphics[width=0.8in]{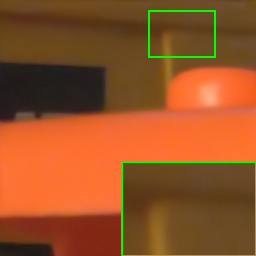}
		35.01dB/0.9270
		\includegraphics[width=0.8in]{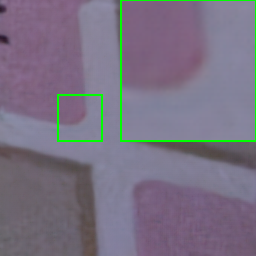}
		36.04dB/0.9145
		\includegraphics[width=0.8in]{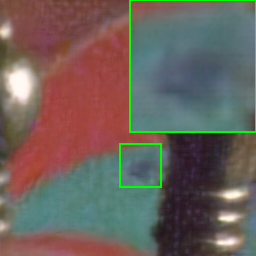}
		28.76dB/0.8002
		\includegraphics[width=0.8in]{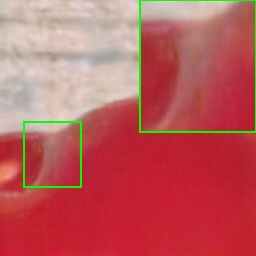}
		28.04dB/0.6693
		\includegraphics[width=0.8in]{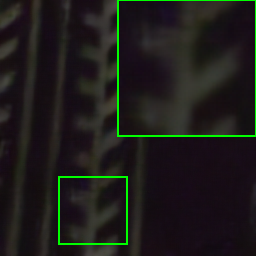}
		35.42/0.9365
		
		\scriptsize{C2N \cite{jang2021c2n}}
	\end{subfigure}
	\begin{subfigure}{0.12\linewidth}
		\centering
		\includegraphics[width=0.8in]{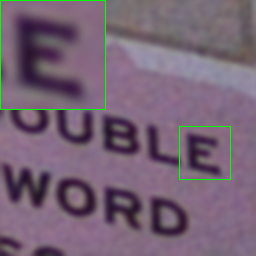}
		31.52dB/0.8935
		\includegraphics[width=0.8in]{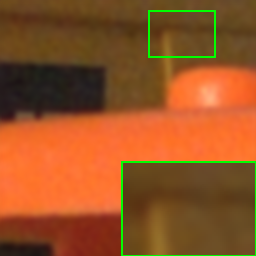}
		34.08dB/0.9244
		\includegraphics[width=0.8in]{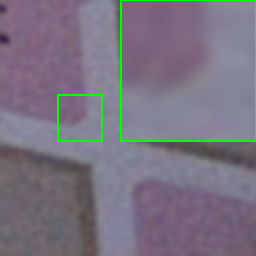}
		34.69dB/0.9130
		\includegraphics[width=0.8in]{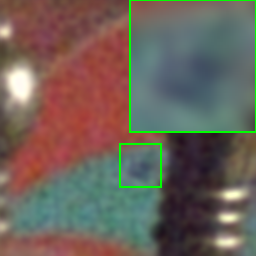}
		28.42dB/0.8201
		\includegraphics[width=0.8in]{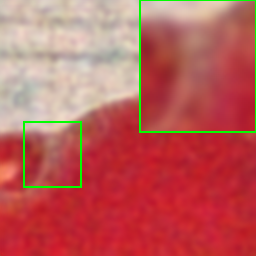}
		27.32dB/0.6727
		\includegraphics[width=0.8in]{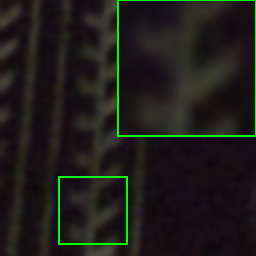}
		34.83/0.9369
		
		\scriptsize{CVF-SID \cite{neshatavar2022cvf}}
	\end{subfigure}
	\begin{subfigure}{0.12\linewidth}
		\centering
		\includegraphics[width=0.8in]{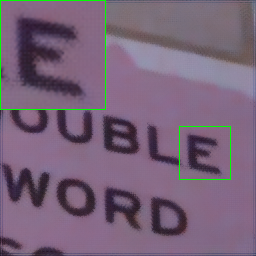}
		32.17dB/0.8988
		\includegraphics[width=0.8in]{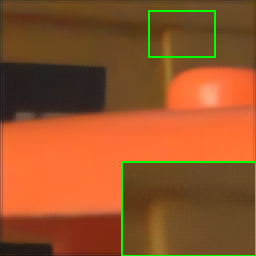}
		33.40dB/0.9224
		\includegraphics[width=0.8in]{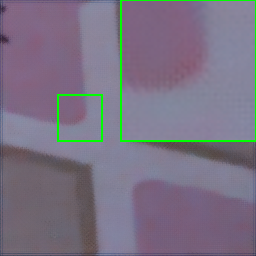}
		34.18dB/0.9002
		\includegraphics[width=0.8in]{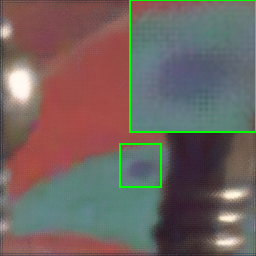}
		30.12dB/0.7868
		\includegraphics[width=0.8in]{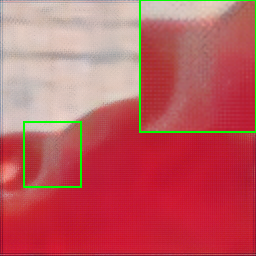}
		26.93dB/0.6002
		\includegraphics[width=0.8in]{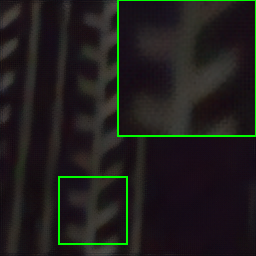}
		36.20/0.9106
		
		\scriptsize{AP-BSN \cite{lee2022ap}}
	\end{subfigure}
	\begin{subfigure}{0.12\linewidth}
		\centering
		\includegraphics[width=0.8in]{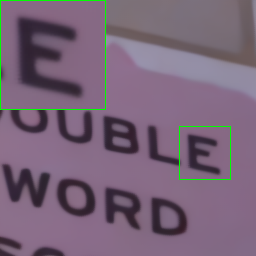}
		36.36dB/0.9350
		\includegraphics[width=0.8in]{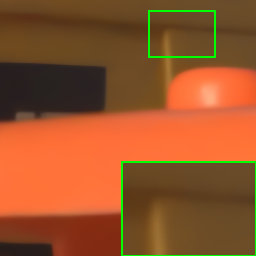}
		38.27dB/0.9430
		\includegraphics[width=0.8in]{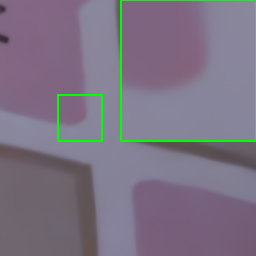}
		38.84dB/0.9377
		\includegraphics[width=0.8in]{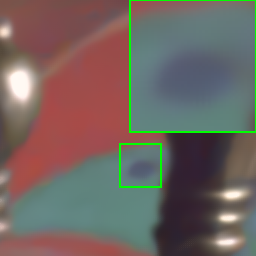}
		32.76dB/0.8586
		\includegraphics[width=0.8in]{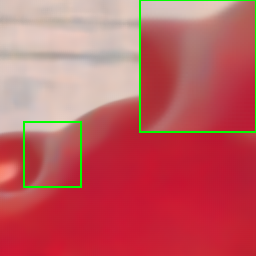}
		29.53dB/0.6850
		\includegraphics[width=0.8in]{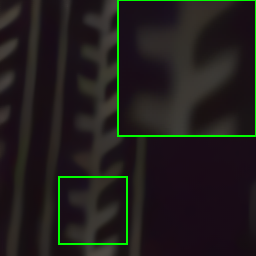}
		38.99dB/0.9537
		
		\scriptsize{\textbf{SDAP (S)(E) (Ours)}} 
	\end{subfigure}
	\vspace{-0.11cm}
	\caption{Visual comparison on the SIDD validation dataset.} 
	\label{sidd}
\end{figure*}

\begin{figure*} \scriptsize
	
	\centering
	\begin{subfigure}{0.12\linewidth}
		\centering
		\includegraphics[width=0.8in]{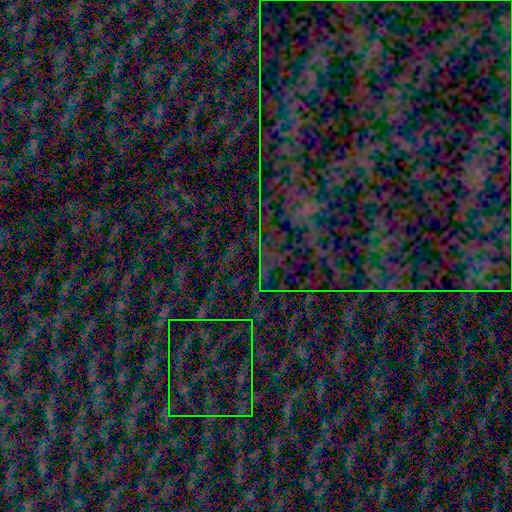}
		PSNR/SSIM
		\includegraphics[width=0.8in]{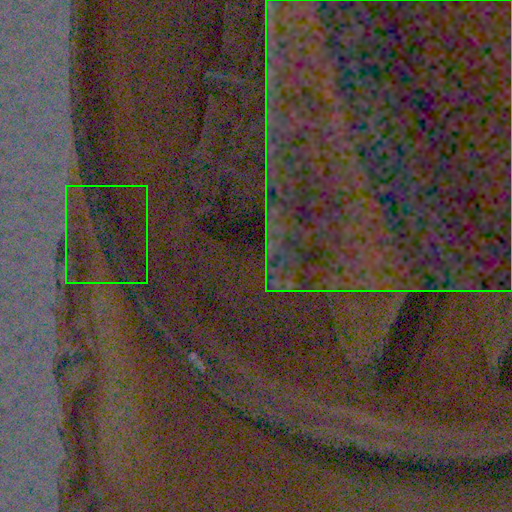}
		PSNR/SSIM
		
		\scriptsize{Noisy}
	\end{subfigure}
	\begin{subfigure}{0.12\linewidth}
		\centering
		\includegraphics[width=0.8in]{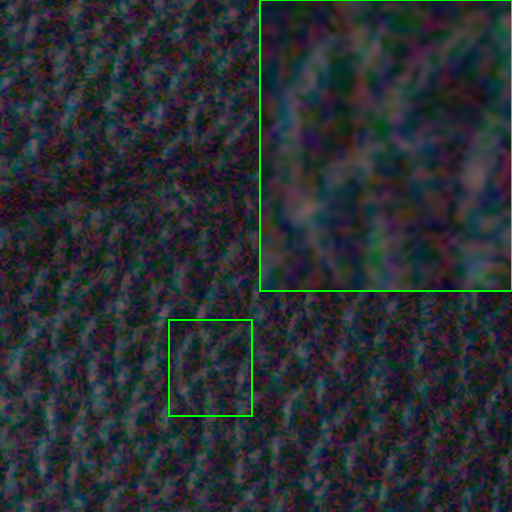}
		25.65dB/0.4580
		\includegraphics[width=0.8in]{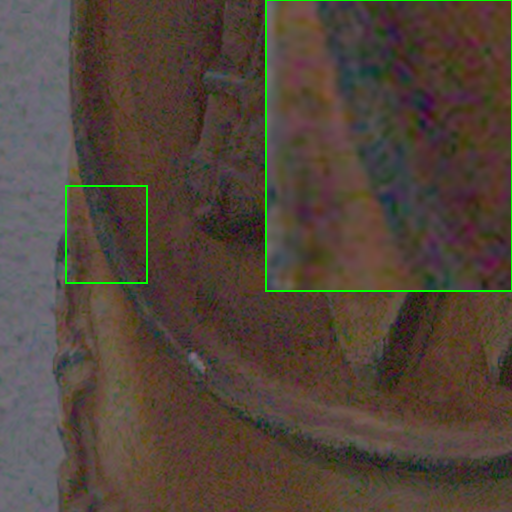}
		29.33dB/0.8175
		
		\scriptsize{BM3D \cite{dabov2007image}}
	\end{subfigure}
	\begin{subfigure}{0.12\linewidth}
		\centering
		\includegraphics[width=0.8in]{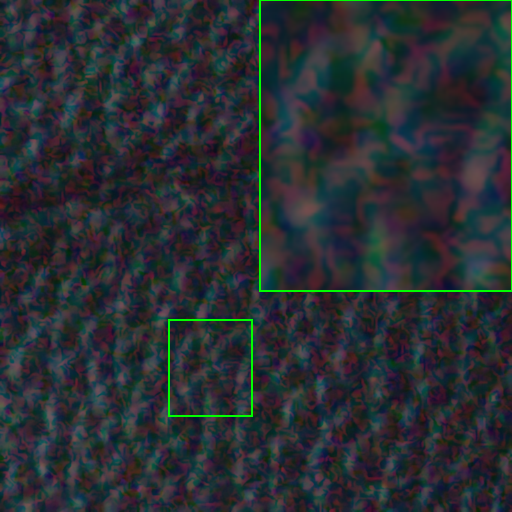}
		25.99dB/0.4848
		\includegraphics[width=0.8in]{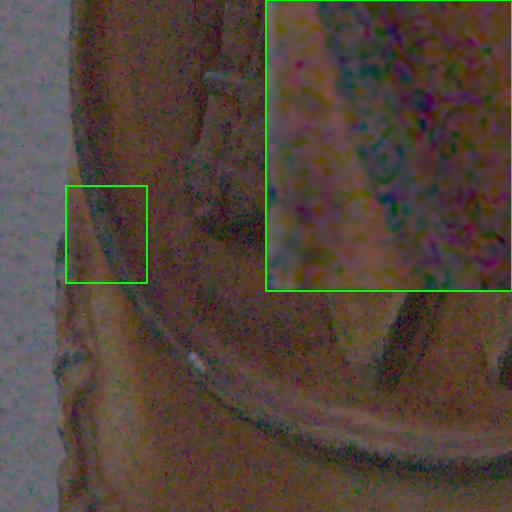}
		26.48dB/0.6884
		
		\scriptsize{DnCNN \cite{zhang2017beyond}}
	\end{subfigure}
	\begin{subfigure}{0.12\linewidth}
		\centering
		\includegraphics[width=0.8in]{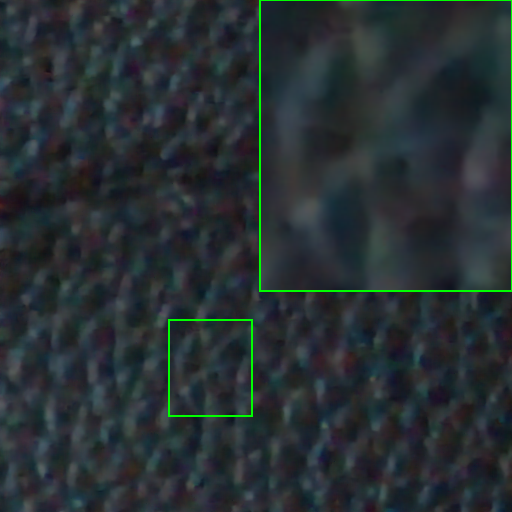}
		31.54dB/0.8254
		\includegraphics[width=0.8in]{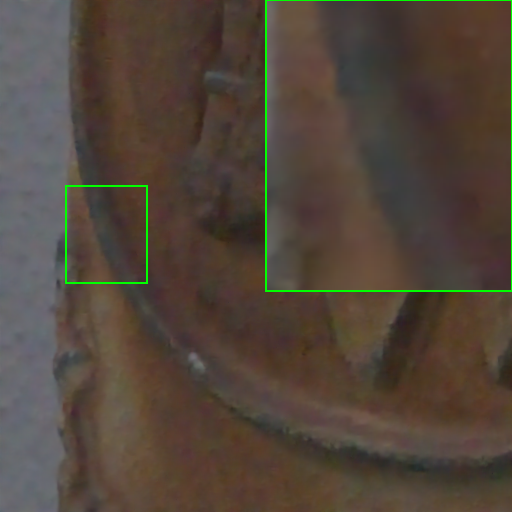}
		35.43dB/0.9469
		
		\scriptsize{CBDNet \cite{guo2019toward}}
	\end{subfigure}
	\begin{subfigure}{0.12\linewidth}
		\centering
		\includegraphics[width=0.8in]{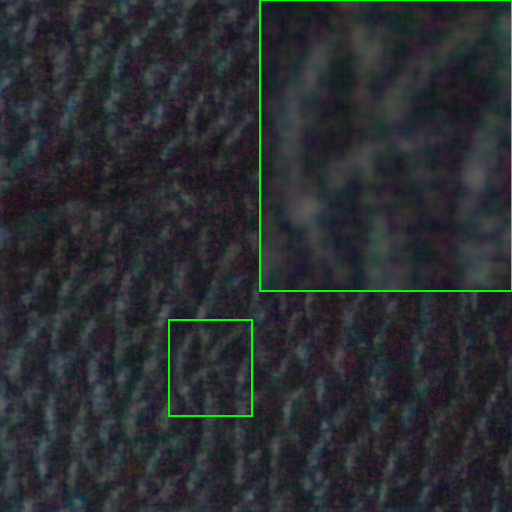}
		29.56dB/0.7064
		\includegraphics[width=0.8in]{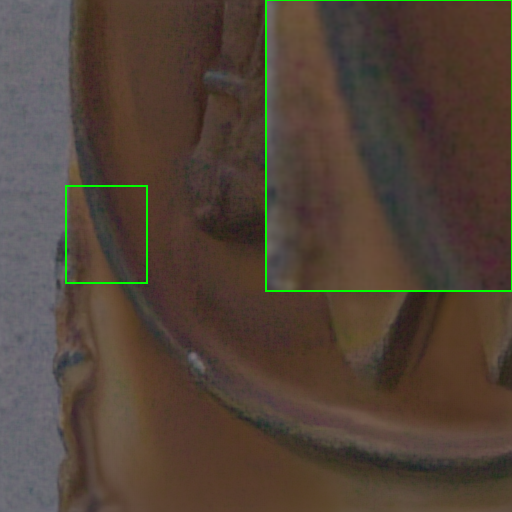}
		33.62dB/0.9432
		
		\scriptsize{C2N \cite{jang2021c2n}}
	\end{subfigure}
	\begin{subfigure}{0.12\linewidth}
		\centering
		\includegraphics[width=0.8in]{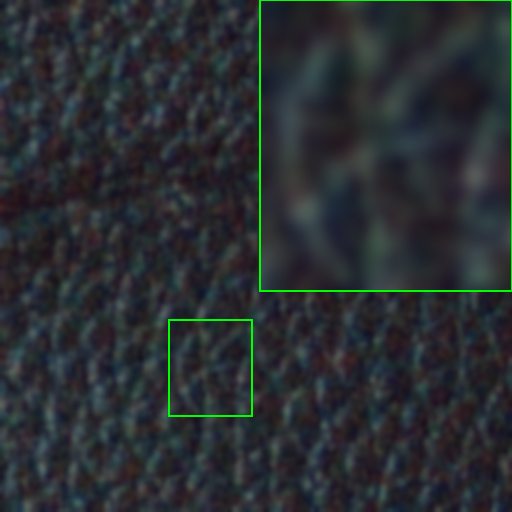}
		31.49dB/0.8098
		\includegraphics[width=0.8in]{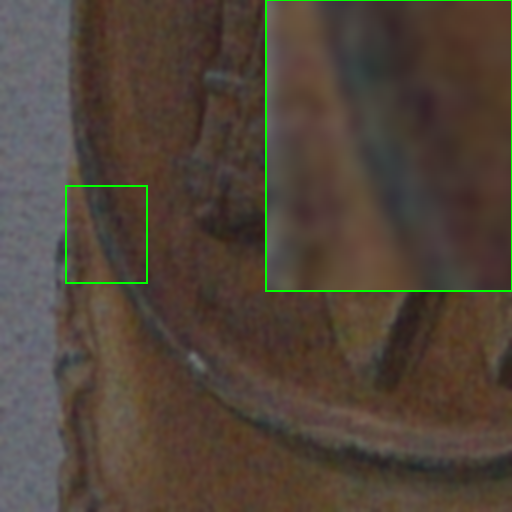}
		34.34dB/0.9468
		
		\scriptsize{CVF-SID \cite{neshatavar2022cvf}}
	\end{subfigure}
	\begin{subfigure}{0.12\linewidth}
		\centering
		\includegraphics[width=0.8in]{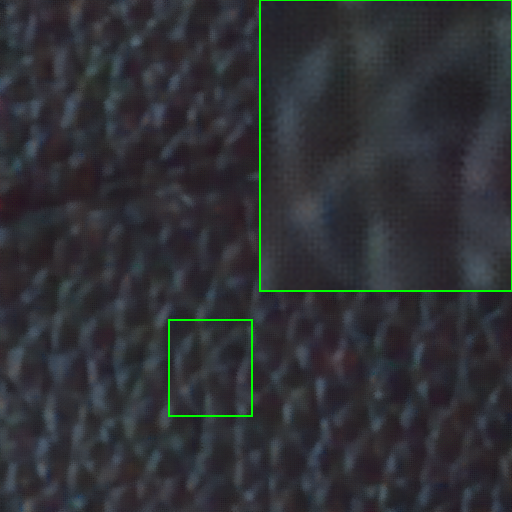}
		31.81dB/0.7812
		\includegraphics[width=0.8in]{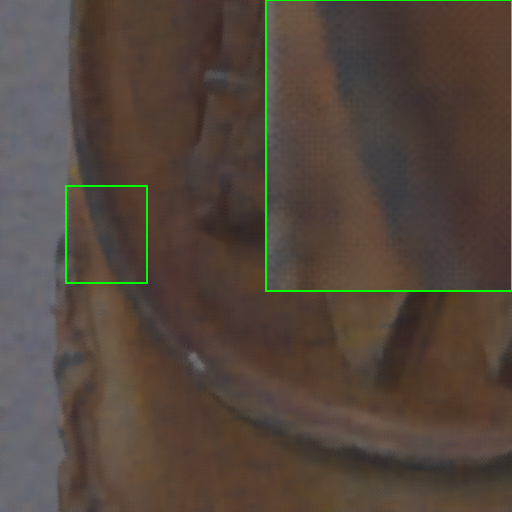}
		35.73dB/0.9448
		
		\scriptsize{AP-BSN \cite{lee2022ap}}
	\end{subfigure}
	\begin{subfigure}{0.12\linewidth}
		\centering
		\includegraphics[width=0.8in]{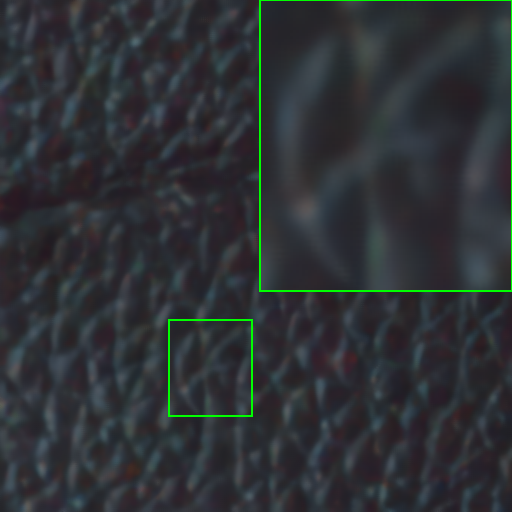}
		32.88dB/0.8337
		\includegraphics[width=0.8in]{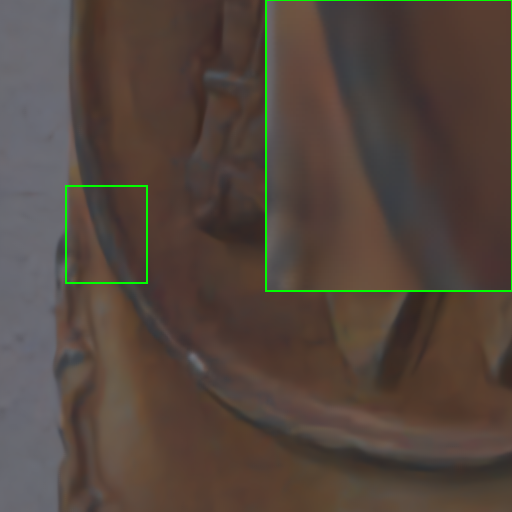}
		36.72dB/0.9570
		
		\scriptsize{\textbf{SDAP (S)(E) (Ours)}} 
	\end{subfigure}
	\vspace{-0.11cm}
	\caption{Denoising examples from DND benchmark.} 
	\vspace{-0.21cm}
	\label{dnd}
\end{figure*}

\section*{S3. Limitations}

In Figure \ref{Limitations}, we illustrate the limitation of our method. Small details in some images are more likely to be considered as noise after being sampled by PD. BSN denoising masks the center pixel, which also causes some image details to be ignored. Since our method uses the PD strategy and uses BSN to denoise the noisy image twice in succession, some small details in the image may be smoothed to some extent. Over-smoothing of details is also a drawback of many unsupervised methods. We still need to continue to work on it, although the results of our method are better than others. For the future work, we hope to obtain better image details using the proposed self-supervised method.

\end{document}